\documentclass[openany, final, los, index, glossary, loa, 10pt]{article}

\usepackage{graphicx}
\usepackage{subcaption}
\usepackage{tikz}

\makeatother

\usepackage{calc}	

\usepackage{tikz}

\usepackage{color}
\usepackage[utf8]{inputenc}
\usepackage{amsmath,amsthm,amsfonts, bm, amssymb} 
\numberwithin{equation}{section} 
\usepackage{url}
\usepackage{chngcntr}
\counterwithin{table}{section}
\usepackage{textcomp}
\usepackage{gensymb}
\usepackage{hyperref}
\hypersetup{pdfborder = {0 0 0}}
\usepackage{rotating} 
\usepackage{booktabs} 

\usepackage{array} 
\usepackage{hhline} 
\usepackage{floatrow} 
\usepackage[margin=1in]{geometry}

\usepackage[T1]{fontenc}
\usepackage{lmodern}

\usepackage[style]{abstract}
\renewcommand{\abstitlestyle}[1]{}

\usepackage[blocks]{authblk}

\renewcommand{\figurename}{Fig. }
\title{A Novel Optimization-Based Collision Avoidance For Autonomous On-Orbit Assembly}

\author{Siavash Tavana \footnote{Ph.D. Student, Department of Aeospace Engineering, Member AIAA.} , Sepideh Faghihi \footnote{Ph.D., Department of Aeospace Engineering, Member AIAA.} , Anton de Ruiter \footnote{Professor and Canada Research Chair in Spacecraft Dynamics and Control, Department of Aerospace Engineering, Associate Fellow AIAA.} , Krishna Dev Kumar \footnote{Professor, Department of Aeropsace Engineering, Associate Fellow AIAA.}}
\affil{Toronto Metropolitan (formerly Ryerson) University, Toronto, ON, M5B 2K3, Canada}

\begin{document}
\date{}
\maketitle

\begin{abstract}
The collision avoidance constraints are prominent as non-convex, non-differen-tiable, and challenging when defined in optimization-based motion planning problems. To overcome these issues, this paper presents a novel non-conservative collision avoidance technique using the notion of convex optimization to establish the distance between robotic spacecraft and space structures for autonomous on-orbit assembly operations. The proposed technique defines each ellipsoidal- and polyhedral-shaped object as the union of convex compact sets; each represented non-conservatively by a real-valued convex function. Then, the functions are introduced as a set of constraints to a convex optimization problem to produce a new set of differentiable constraints resulting from the optimality conditions. These new constraints are later fed into an optimal control problem to enforce collision avoidance where the motion planning for the autonomous on-orbit assembly takes place. Numerical experiments for two assembly scenarios in tight environments are presented to demonstrate the capability and effectiveness of the proposed technique. The results show that this framework leads to optimal non-conservative trajectories for robotic spacecraft in tight environments. Although developed for autonomous on-orbit assembly, this technique could be used for any generic motion planning problem where collision avoidance is crucial.
\end{abstract}

~\\


\section{Introduction} \label{intro}

Autonomous on-orbit assembly (AOA) is gaining significant attention in the space exploration community. The need for building large, complex space structures in deep space and maintaining them, as proposed in the International Space Exploration Coordination Group (ISECG) roadmap \cite{ISECG2013} and a NASA study \cite{belvin2016space}, has recently attracted researchers to develop algorithms and techniques for autonomous on-orbit assembly along with other on-orbit proximity operations, such as inspection, refueling, and repair \cite{faghihi2023multiple,nakka2022information,davis2019orbit,guang2018attitude}. On the one hand, the limitations on the size and weight of the launch vehicles necessitate on-orbit assembly. On the other hand, the issues due to long distances between the operation location in deep space and the control ground station demand other alternatives to approach such missions. A proposed solution is to do assembly in low earth orbit, before transportation to deep space locations. Such assembly could be done with ground in the loop, or by astronauts. There may still be a limitation in the size of assembled structures that can be transported to deep-space along with the need for Extra-Vehicular Astronaut Activity (EVA) which is dangerous and expensive in nature. In addition, an active area of investigation is in-situ resource extraction, and subsequent in-space manufacture. Deep-space structures that are manufactured from materials mined remotely will require autonomous assembly. 

AOA, in general, consists of three main components that need to be considered in order to perform a complete AOA operation: 1. task allocation system where the planning and scheduling of the assembly tasks are designed either by human intervention or autonomously, 2. maneuvering the structural modules to the assembly point where the required guidance, navigation, and control techniques should be developed, and 3. utilizing local robotic manipulators to assemble the structural modules where the robotic motion planning techniques are of the essence. Depending on the application, AOA could be performed in two methods: 1. The first method involves a collection of robotic spacecraft cooperatively maneuvering structural modules \cite{komendera2017initial}, while 2. the second method is to employ the structural modules equipped with control means, enabling them to maneuver themselves to the assembly and docking points, where robotic manipulations would not be required. \cite{badawy2008orbit}.

Task allocation has been addressed in general assembly applications \cite{hartmann2022long}; however, it mainly involves human supervision through teleoperation in in-space applications \cite{nanjangud2019towards, rognant2019autonomous}. These techniques would not be viable for longer distances and more complex missions in deep space. Nevertheless, some researchers have tried to shift this paradigm toward full autonomy by introducing a state space to track each module and employing graph search methods for planning and evaluating the assembly schedule \cite{rodriguez2021autonomous}. One step further, the authors of \cite{moser2022autonomous} utilized stochastic problem formulation and mixed integer programming to generate an optimal task scheduling for a fully autonomous AOA operation.

Regarding the robotic manipulation in the final phase, much research has been done on the localizing, grasping, and manipulation \cite{komendera2017initial}. For instance, the work in \cite{lu2020experimental} utilizes a fixed manipulator to grasp a structural module and assemble it to the fixed central structure using iterative learning control and an economical camera for optical measurement. In addition, many experimental works have been done for truss deployment and modular lattice on the ground for feasibility purposes \cite{karumanchi2018payload, jenett2017bill}. Although these are valuable advances toward AOA development, the analysis of free-floating base manipulators is of great importance as this is the case for in-space assembly operations. The authors of \cite{virgili2017laboratory} demonstrated the capture of a resident free-floating object using a free-floating robotic spacecraft with a single manipulator utilizing a resolved motion rate controller, while the authors in \cite{jayakody2016robust} have studied the effects of the manipulator motion on the free-floating base by decoupling the highly nonlinear dynamics using multiple input/output system model. Moreover, on the pre-grasping side, a nonlinear model predictive control was used to grasp a spinning object using a dual-arm, free-floating spacecraft \cite{wang2020coordinated}, while in the post-grasping side, researchers in \cite{stolfi2020two} developed a Jacobian transpose and proportional-derivative controller to manipulate a passive object in the presence the grasp constraints.

Finally, in terms of maneuvering the structural modules, most work in the literature are devoted to the second method of AOA, where modules maneuver themselves with built-in control means, rather than the first method, where modules are maneuvered by robotic spacecraft. For example, a distributed control law with disturbance observer was proposed in \cite{chen2018autonomous} to assemble a fleet of self-maneuvered, flexible modules to a large structure, while generating optimal trajectories for docking was studied in \cite{jewison2018probabilistic} under uncertainties with a probabilistic optimal level of performance to account for mission safety and accuracy. Additionally, some works were devoted to experimentally evaluating guidance, navigation, and control techniques developed for AOA purposes \cite{wilde2016experimental, pellegrini2010spacecraft}. On the contrary, the literature on maneuvering modules using robotic spacecraft could be much more extensive, and there is much to be explored and developed. From that, the work in \cite{rekleitis2015orbit} can be mentioned where a model-based controller was proposed to handle a sizeable passive object under two different contact-type scenarios (point and firm contacts) using multiple robotic servicers. Simulations have shown that fuel consumption has decreased, and desired positioning accuracy was obtained.

The authors believe that what hindered the researchers from turning their focus to the autonomous maneuvering of structural modules is the lack of a robust and non-conservative collision avoidance. At the same time, this paradigm could be a niche technique for cost-effective, accurate, and efficient AOA operations. Moreover, it is noteworthy that even structures assembled by self-maneuvered modules need to be maneuvered by a set of robotic spacecraft at some point throughout the operation to a desired location when they become larger after a few steps. The collision avoidance constraints are prominent to be non-convex, non-differentiable, and challenging to deal with in general when defined in optimization-based motion planning problems. What makes the collision avoidance problem in AOA different from other space docking and rendevous applications is the presence of plenty of structural modules and robotic spacecraft with complex shapes that are attached and closely moving simultaneously so that the assembly operation can be completed. As a result, any proposed collision avoidance technique must consider the full shape of the active components according to the nature of the mission.


Optimization-based trajectory generation is a core technique in motion planning problems for aerospace vehicles. This is mainly because of the low dimensionality of problems in the field. Collision avoidance constraint is an inseparable matter in optimization-based methods. Since the collision avoidance constraint is inherently non-convex and non-differentiable in optimal control problems, many works have been done to formulate it as a convex and differentiable constraint. The vast majority of these works consider the controlled vehicles and obstacles as point-mass objects due to their formulation simplicity. For instance, \cite{leomanni2022variable} and \cite{foust2020autonomous} define a spherical boundary with a certain radius for each point-mass spacecraft and examine the distance between each two spacecraft to ensure that they maintain the minimum distance defined by the radius. Others, such as \cite{adhikari2020online} and \cite{dong2017safety}, consider the obstacles as rigid bodies and use an artificial potential function in the cost functional to deter the controlled object from approaching the obstacles. To improve the situation, some encapsulate the controlled vehicle and obstacles by ellipsoids and then enforce collision avoidance by this approximation using a rotating hyperplane attached to the obstacle's ellipsoid, which rotates at a constant rate to guide away the spacecraft, or a dual hyperplane method, which fixes two stationary hyperplanes at two points on the obstacle's keep-out boundary \cite{zagaris2018model}. These ellipsoids are sometimes utilized to check an inequality constraint defined by the ellipsoids' equations to check whether a point is within the ellipsoid \cite{jewison2015model, park2017nonlinear}.  

In the case of docking missions, many have employed the line-of-sight (LoS) cone constraint to prevent a collision between the approaching and target spacecraft. Hyperplanes are used to define the LoS cones, and a number of inequalities are introduced, depending on the number of hyperplanes used to form the LoS cone, to maintain the approaching spacecraft in the desired corridor \cite{di2012model, li2018model, weiss2015model}. Finally, \cite{virgili2017convex} uses the signed distance function to define the distance between the two objects, then approximates the signed distance function by linearizing the function, and ultimately, solves it using sequential convex programming. 


In the robotic context, similar evolution can be seen in the literature. While some utilized barrier functions in the cost functional \cite{obeid2018barrier, wu2019control}, others have preferred to define the obstacle avoidance constraint directly in the optimal control problem. The first preliminary approaches introduced the robots' and obstacles' shapes as points, lines, paraboloids, and ellipsoids \cite{blackmore2011chance, brown2019coordinating, rosolia2016autonomous}. This way of shape expression results in analytical and differentiable constraints in the optimization problem. These methods have been improved by introducing polytopes in the constraints, which was less conservative but still had an implicit and non-analytical expression for distance function \cite{grossmann2002review}. In addition, the non-differentiability of the distance function with respect to the robot states is another challenge in including the polytopes \cite{deits2015efficient}. To address these issues, some solutions have been proposed. For example, Deiths and Tedrake \cite{deits2015efficient} use mixed-integer convex optimization to solve for the distance function; however, it is not suitable for nonlinear systems.

On the other hand, the signed distance function is a popular way to define the distance constraint in an optimal control problem. Still, the signed distance function is not straightforward to include in the problem, as there is no explicit representation of it with respect to the decision variables. The authors of \cite{schulman2014motion} deal with the signed distance function directly by sequentially linearizing the signed distance to estimate its gradients. This method can fail in degenerate situations where a face of the controlled object becomes parallel to the obstacle's. In addition, the approximation error of linearization cannot be established. To avoid these problems, The authors of \cite{zhang2020optimization} propose a duality-based method to solve for the signed distance function in order to generate a new set of differentiable constraints in the optimization problem. Ultimately, the notion of discrete control barrier function has been used as an alternative to applying the collision avoidance constraint between two polytopes in the optimization problem by \cite{thirugnanam2022safety}. Although the last two methods are interesting, they consider polytopes only as their general shapes, and their methods drastically increase the problem's dimension by introducing new variables, especially in the case of full-dimensional controlled objects.

As a result, although many methods have been proposed in the literature to deal with collision avoidance constraints in optimization-based trajectory generation methods, they at least have one of the following limitations, making them inappropriate for AOA operations.

\begin{enumerate}

\item Current methods concentrate on point-mass controlled vehicles, which makes them unsuitable for applications that deal with full-dimensional bodies \cite{morgan2014model, blackmore2011chance}.

\item The methods developed for full-dimensional controlled vehicles in aerospace literature are too conservative and not applicable to AOA purposes where proximity movements are the core issue \cite{virgili2017convex, weiss2015model}.

\item Other techniques proposed for full-dimensional controlled vehicles moving in similar environments as AOA's, such as those in the robotic literature, do not account for the considerations of space applications \cite{thirugnanam2022safety, zhang2020optimization}.   

\end{enumerate}

Consequently, the necessity of proposing a collision avoidance technique that is not conservative in any manner, is designed for full-dimensional spacecraft and obstacles, and is comprehensive to incorporate the space robotics applications, as well as other applications, is evident. As a result, this paper, in general, proposes a novel collision avoidance for a collection of full-dimensional controlled vehicles that move in a general n-dimensional space to overcome the challenges in the existing formulations mentioned above. In summary, the specific contributions of this paper are:

\begin{enumerate}

\item A novel approach is proposed to reformulate the collision avoidance problem by precisely modelling complex-shaped, full-dimensional controlled vehicles and obstacles when they are made of the union of a finite collection of convex compact sets using a series of differentiable convex functions. These functions are later used as a series of constraints in a convex optimization problem to establish the minimum distance between any two convex sets. The optimality condition of this optimization problem forms a new set of differentiable constraints guaranteeing the generation of a collision-free trajectory in any optimal control problem.

\item The efficacy of the proposed method is demonstrated in an autonomous on-orbit assembly operation in deep space, where the working environment is tight for the robotic spacecraft that performs the assembly procedure. A pesudo-specteral optimal control method \cite{ross2015primer} is utilized as a solver to show that the proposed method can generate optimal trajectories in tight environments with multiple active components present.

\end{enumerate}

The paper's organization is as follows. Section \ref{prob-statement} describes the general optimal control problem with collision avoidance constraints. The collision avoidance constraints are reformulated for both a point-mass and full-dimensional controlled vehicle in Section \ref{reform-colavoid-const} by modeling objects using differentiable convex functions. Section \ref{app-aoa} presents numerical simulations illustrating the effectiveness of the novel method in an autonomous on-orbit assembly mission. Ultimately, conclusions are drawn in Section \ref{conclusions}. \\


\section{Problem Statement} \label{prob-statement}


\subsection{Dynamics, Constraints, Performance Index} \label{dyn-const-pi}

The problem at hand is to generate an optimal trajectory using optimal control problem formulation considering the general mission constraints on the states and control inputs, as well as collision avoidance constraints. In general, the dynamics of the moving vehicle can be written in the form

\begin{equation} \label{dyn} \dot{\boldsymbol{x}} = \boldsymbol{f}(\boldsymbol{x}, \boldsymbol{u}, t) \end{equation}

\noindent where $\boldsymbol{x} \in \mathbb{R}^{n_x}$ is the state, $\boldsymbol{u} \in \mathbb{R}^{n_u}$ is the control input, and $\boldsymbol{f} : \mathbb{R}^{n_x} \times \mathbb{R}^{n_u} \times \mathbb{R} \rightarrow \mathbb{R}^{n_x}$ is the dynamics of the system. The system's state consists of the position and velocity information of the controlled vehicle when it is a point-mass object. When it is a full-dimensional body, the state represents position, orientation, translational and angular velocities (or angular rates). All constraints on the state and control input except the collision avoidance constraint can be written as

\begin{equation} \label{gen-const} \boldsymbol{g}(\boldsymbol{x}, \boldsymbol{u}, t) \leq 0 \end{equation}

\noindent where $\boldsymbol{g} : \mathbb{R}^{n_x} \times \mathbb{R}^{n_u} \rightarrow \mathbb{R}^{n_g}$. $n_g$ determines the number of elementwise inequality constraints on the problem aside from the dynamics. Analogous to any optimal trajectory generation, we want to optimize some performance index $h(\boldsymbol{x},\boldsymbol{u},t)$, where $h : \mathbb{R}^{n_x} \times \mathbb{R}^{n_u} \times \mathbb{R} \rightarrow \mathbb{R}$, over the period of $\begin{bmatrix} t_0 & t_f \end{bmatrix}$, where $t_0$ and $t_f$ are the initial and terminal times, respectively. This performance index can be presented in terms of the cost functional as

\begin{equation} \label{cost-func} J = \int_{t_0}^{t_f} {h(\boldsymbol{x}, \boldsymbol{u}, t) dt} \end{equation}


\subsection{Collision Avoidance Constraint} \label{nonconv-nondiff-colavoid}

The collision avoidance constraint between the controlled vehicle and obstacles can be written in its original form as

\begin{equation} \label{org-colavoid} S(\boldsymbol{r}_I^{s}, \boldsymbol{C_{sI}}) \cap O_i(t) = \emptyset \quad \mbox{for } i = 1, ..., n_o  \end{equation}
\vspace{0.5em}

\noindent where $S$ and each $O_i$ are the sets representing the controlled vehicle and obstacles, respectively. The controlled vehicle and obstacles are assumed to be constructible by the union of a finite collection of convex compact sets; therefore, sets $S$ and $O_i$s can all be broken down into a finite collection of convex sets. The collision avoidance proposed in this paper will take advantage of this property later. Moreover, if the controlled vehicle was point-mass, set S would be only the function of $\boldsymbol{r}_I^s$, which is the position vector of the point of interest $s$ on the controlled vehicle with respect to the inertial reference frame, defined in the problem. In this case, set $S$ would be a singleton with its only member point $s$. Otherwise, set $S$ would be the function of the position vector $\boldsymbol{r}_I^s$ and the rotation matrix $\boldsymbol{C}_{BI}$ of the body reference frame, fixed at the point of interest $s$, with respect to the inertial frame. As seen in Eq. (\ref{org-colavoid}), the obstacles are the function of time, meaning they can move in space.


\subsection{Optimal Control Problem Formulation} \label{ocp-form}

Having the dynamics, constraints, and performance index defined, the optimal control problem can be formulated using Eqs. (\ref{dyn} - \ref{org-colavoid}) as

\begin{subequations} \begin{eqnarray}  \underset{x, u, t}{\mathrm{min}} &&  J = \int_{t_0}^{t_f} {h(\boldsymbol{x}, \boldsymbol{u}, t) dt} \label{ocp-min} \\ \mbox{subject to: } && \boldsymbol{x}_0 = \boldsymbol{x}_I, \; \boldsymbol{x}_f = \boldsymbol{x}_T \label{ocp-init-term-cond} \\ && \dot{\boldsymbol{x}} = \boldsymbol{f}(\boldsymbol{x}, \boldsymbol{u}, t) \label{ocp-dyn} \\ && \boldsymbol{g}(\boldsymbol{x}, \boldsymbol{u}, t) \leq 0 \label{ocp-const} \\ && S(\boldsymbol{r}_I^{s}, \boldsymbol{C_{sI}}) \cap O_i(t) = \emptyset \quad \mbox{for } i = 1, ..., n_o \label{ocp-colavoid} \end{eqnarray} \label{ocp} \end{subequations}

The optimal control problem, as formulated above, would be challenging to solve, even in the case where the objective function Eq. (\ref{ocp-min}), dynamics Eq. (\ref{ocp-dyn}), and state and control input constraints as in Eq. (\ref{ocp-const}) are all convex and differentiable, because the collision avoidance constraint is non-convex and non-differentiable as written in Eq. (\ref{ocp-colavoid}). The following section proposes a novel reformulation of the collision avoidance constraint to address this issue, resulting in continuous and differentiable constraints. \\


\section{Reformulation of Collision Avoidance Constraint} \label{reform-colavoid-const}

To be able to use the gradient- and Hessian-based optimization solvers, all the functions in the optimal control problem must be continuous and differentiable with respect to the optimization variables. To reformulate collision avoidance, the controlled vehicle and obstacles are first modeled using real-valued, differentiable, convex functions. These functions are fed into a convex optimization problem as a set of constraints to generate a new set of constraints from the optimality conditions of such an optimization, establishing the minimum distance between any two convex sets defined by those functions. This results in a new collection of differentiable constraints in the optimal control problem, which enforces collision avoidance while solving for optimal trajectories.


\subsection{Controlled Vehicle And Obstacles Modeling}\label{sc-st-models}

Almost all objects can be divided into two categories of polyhedral and ellipsoidal shapes when it comes to involving objects in aerial, space, ground, and marine applications, where they can be constructed by the union of a finite collection of convex compact sets. To begin with, the distance between any two non-empty, compact, convex sets $C$ and $D$ in an n-dimensional vector space with the differentiable norm $\lVert . \rVert$ can be formulated as \cite{Boyd2004convex}

\begin{subequations} \begin{eqnarray}  \mathrm{min} && \lVert \boldsymbol{w} - \boldsymbol{p} \rVert \\ \mbox{subject to:} && f_{C_i}(\boldsymbol{w}) \leq 0, \quad i = 1, ..., m_1 \\  && f_{D_j}(\boldsymbol{p}) \leq 0, \quad j = 1, ..., m_2 ,\end{eqnarray} \label{dist-opt} \end{subequations}

\noindent where each set can be defined by a convex inequlaity as

\begin{equation} \label{set-def} C = \{  \boldsymbol{w}: f_{C_i}(\boldsymbol{w}) \leq 0, \; i = 1, ..., m_1 \} \quad \quad D = \{  \boldsymbol{p}: f_{D_j}(\boldsymbol{p}) \leq 0, \; j = 1, ..., m_2 \} ,\end{equation}

\vspace{1em}

Sets' closedness is required because the solution to this optimization problem is always on the boundary of the convex set. Therefore, the boundary should be included in the feasible set of the optimization problem. If that were not the case, then the solution to the optimization problem would not be in the feasible set. In addition, since every structure is bounded, a closed convex set representing them is also compact. Now in order to take advantage of the above convex optimization properties to find the distance between two convex sets, the p-norm is utilized to define convex sets. To begin with, set $ E $ will be defined as an ellipsoid first and then as a polyhedron just to prove the proposed mathematical properties for both shape categories. The following results apply to any combination of polyhedral and ellipsoidal shape pairs, i.e., ellipsoid-ellipsoid, polyhedron-polyhedron, and polyhedron-ellipsoid. On the other hand, to resolve the orientation and translation of the moving sets throughout time with respect to the inertial coordinate system, the point $\boldsymbol{z}$ on set $E$, is defined with respect to its corresponding body-fixed reference frame attached to the center of the set. In contrast, $\boldsymbol{z}_I$, defined with respect to the inertial coordinate frame, is a function of the attitude matrix and position vector. Therefore, the points are correlated as  

\vspace{1em}
\begin{equation}\label{rot-trans} \boldsymbol{\mathcal{F}}_I^T \boldsymbol{p}_I = \boldsymbol{\mathcal{F}}_{I}^T \boldsymbol{C}_{IB}(t) \boldsymbol{p} + \boldsymbol{\mathcal{F}}_{I}^T \boldsymbol{d}_{I}(t), \quad \boldsymbol{p} \in E \end{equation}
\vspace{1em}

\noindent where $\boldsymbol{C}_{IB}(t)$ and $\boldsymbol{d}_{I} (t)$ are the rotation matrix and the translation vector of the body-fixed frame with respect to the inertial coordinate system. In what follows, the functions used to describe the sets are all defined with respect to their corresponding body-fixed reference frame, and then the defined sets can be obtained in the inertial coordinate frame using Eq. (\ref{rot-trans}) as

\vspace{1em}
\begin{equation}  E_I(t) = \boldsymbol{C}_{IB}(t) E + \boldsymbol{d}_{I}(t) \end{equation}
\vspace{0.5em}

Now without loss of generality, let set $E$ be any shape in $\mathbb{R}^n$ that can be defined by the real-valued, differentiable function $f_C: \mathbb{R}^n \rightarrow \mathbb{R}$ as

\begin{equation} \label{c-func}  f_E(\boldsymbol{p}) = (\sum_{i = 1}^{n}{f_{E_i}(p_i)}) - l^{P_{n+1}}, \quad \mbox{where } f_{E_i}(p_i) = ( \frac{p_i}{a_i})^{P_i}, \; i=1,..., n, \end{equation}

\noindent where $\boldsymbol{p} = (p_1, p_2, ..., p_n) \in \mathbb{R}^n$, and the center of the shape is defined on the origin of the body-fixed frame. Also, $\boldsymbol{a} = (a_1, ..., a_n) \in \mathbb{R}^n$ is the shape factor, and $l \in \mathbb{R}$. In addition, $P_i$ for $i = 1, ..., n$ are the exponents and, without loss of generality, assumed to be even for continuity and differentiability purposes. Function $f_{E}(\boldsymbol{p})$ maps the points in $\mathbb{R}^n$ to define set $E$. According to $f_E(\boldsymbol{p})$, if $f_E(\boldsymbol{p}) < 0 $, that means point $\boldsymbol{p}$ is an interior point, if $f_E(\boldsymbol{p}) = 0$, point $\boldsymbol{p}$ is a boundary point, and if $f_E(\boldsymbol{p}) > 0$, point $\boldsymbol{p}$ is an exterior point of set $E$. Therefore, set $E$ can be written as

\begin{equation} \label{setC} E = \{  \boldsymbol{p}: f_E(\boldsymbol{p}) \leq 0  \}  \end{equation}
\vspace{0.5em}

On the other hand, any set $E$ with polyhedral shape in $\mathbb{R}^n$ can be defined using a real-valued, differentiable function $f_E: \mathbb{R}^n \rightarrow \mathbb{R}$ as

\begin{equation} \label{d-func}  \boldsymbol{f}_{E_{ex}}(\boldsymbol{p}) = \boldsymbol{A} \boldsymbol{p} - \boldsymbol{b} \end{equation}
\vspace{0.5em}

\noindent where $\boldsymbol{p} \in \mathbb{R}^n$, $\boldsymbol{A} \in \mathbb{R}^{m \times n}$ is the shape matrix, and $\boldsymbol{b} \in \mathbb{R}^m$. Based on $f_E(\boldsymbol{p})$, point $\boldsymbol{p}$ is an interior point of set $E$ if $f_E(\boldsymbol{p}) < 0$, a boundary point if any of the inequalities is equal to zero, and an exterior point if at least one inequality is greater than zero. Consequently, set $E$ in its exact form can be defined as

\vspace{0.5em}
\begin{equation} \label{setD-ex} E_{ex} = \{  \boldsymbol{p}: \boldsymbol{f}_{E_{ex}}(\boldsymbol{p}) \leq 0   \}  \end{equation}
\vspace{0.5em}

As will be seen later in the paper, the way that function $\boldsymbol{f}_{E_{ex}}$ and set $E_{ex}$ have been modeled would increase the optimal control problem dimension by introducing the Lagrange multipliers as $\boldsymbol{\lambda} \in \mathbb{R}^m$, since set $E_{ex}$ consists of \emph{m} elementwise inequalities. To improve this, function $\boldsymbol{f}_{E_{ex}}$ is redefined as

\vspace{0.5em}
\begin{equation} \label{d-func2} f_{E_{new}}(\boldsymbol{p}) = \mathrm{max}\{  f_{E_j}(\boldsymbol{p})  \}, \quad \mbox{where } f_{E_j}(\boldsymbol{p}) = A_j \boldsymbol{p} - b_j \; \mbox{for } j = 1, ..., m \end{equation}
\vspace{0.5em}

\noindent where $A_j$s as in $\boldsymbol{A} = \begin{bmatrix} A_1 & \hdots & A_m \end{bmatrix}^T$ are the rows of matrix $A$ and $\boldsymbol{b} = \begin{bmatrix} b_1 & \hdots & b_m \end{bmatrix}^T$. \\

$f_{E_{new}}$ is non-differentiable as written in Eq. (\ref{d-func2}), which is problematic when used in gradient-based solvers. To address that, we will use p-norm representation to make $f_{E_{new}}(\boldsymbol{p})$ differentiable. According to the p-infinity norm definition, it singles out the maximum of the absolute value of the terms \cite{clarke2013functional}. Since $f_{E_j}$s have negative values, using the p-infinity norm directly would not work here; therefore, we take the exponential of $f_{E_j}$s to map all values to a range between zero and one, apply the p-infinity norm to extract the maximum value, and then take the natural logarithm of the terms to map them back to the original range. Consequently, we modify $f_{E_{new}}$ to get

\vspace{0.5em}
\begin{equation}\label{d-func3} f_{E_{app}}(\boldsymbol{p}) = ln[(\sum_{j = 1}^{m} {e^{f_{E_j}(\boldsymbol{p})  Q_j}})^{\frac{1}{Q_{m+1}}}], \quad Q_j, Q_{m+1} >> 0, \; j= 1,...,m \end{equation}
\vspace{0.5em}

Since infinity is mathematically challenging, $Q_j$s for $j=1, ..., m+1$ cannot be infinity; consequently, large numbers should be selected for $Q_j$s depending on the application, where the accuracy of the approximation would be set. Now with this approximation, $f_{E_{app}}$, as defined in Eq. (\ref{d-func3}), does not exactly model the sharpness of the polyhedron vertices where the approximation takes effect. To provide more control over the approximation, the logarithm and exponent base can be adjusted when $Q_j$s are selected to improve the modeling accuracy due to smoothing approximation at vertices. Ultimately, $f_{E_{app}}$ in its complete and final form describes set $E$ with an approximation as

\begin{equation} \label{setD-app} E_{app} = \{  \boldsymbol{p}: f_{E_{app}}(\boldsymbol{p}) \leq 0   \}  \end{equation}

The following results in this paper are proven for both the exact and approximated polyhedron models. It is further the choice of the designer to use the exact or approximated model for high accuracy or fast computation, respectively, depending on the application. Additionally, to model non-ordinary shapes consisting of the intersection of polyhedral and ellipsoidal shapes, the combination of the corresponding functions, as introduced in Eqs. (\ref{c-func}), (\ref{d-func}), and (\ref{d-func3}) can be used in the set definition in Eq. (\ref{set-def}) to define such sets. Now, all that is left to show is the convexity of $f_E$, $f_{E_{app}}$, and $f_{E_{ex}}$.

\emph{Proposition 1:} Functions $f_E$, $f_{E_{ex}}$, and $f_{E_{app}}$ are convex.

\emph{Proof:} Since $f_{E_i}$s, as in Eq. (\ref{c-func}) are binomials with even exponents, their second derivatives have an even exponent as well, resulting in a non-negative second derivative. Therefore, $f_{E_i}$s are convex \cite{Boyd2004convex}. On the other hand, function $f_E$, as seen in Eqs. (\ref{c-func}), is a linear combination of $f_{E_i}$ for $i = 1,...n$. Since each $f_{E_i}$ is convex, $f_E$ becomes convex. The convexity of $f_{E_{ex}}$ directly follows from its definition as it consists of \emph{m} elementwise linear inequalities \cite{Boyd2004convex}. Regarding the convexity of $f_{E_{app}}$, we use this thought process that the minimum of $f_{E_{app}}$ is reached at the center of the polyhedron which is a negative value, and therefore, moving in any direction would increase the function value up to the boundary of the polyhedron where $f_{E_{app}}$ reaches zero based on its definition in Eq. (\ref{setD-app}). Continuing this process would result in increasing positive values for the function as the evaluated point gets far away from the polyhedron. On the other hand, since $f_{E_{app}}$ is continuous on set $E_{app}$ and has a minimum on it according to its definition; hence, its second derivative is non-negative, and therefore, $f_{E_{app}}$ is convex \cite{Boyd2004convex}.  \hfill $\square$

As the result of \emph{Proposition 1}, the optimization problem in Eq. (\ref{dist-opt}) becomes a convex optimization problem, for which the strong duality holds since sets $C$ and $D$, defined based on Eqs. (\ref{setC}), (\ref{setD-ex}), and (\ref{setD-app}), have a non-empty interior \cite[p. 226]{Boyd2004convex}. \\


\subsection{Collision Avoidance For Point-Mass Controlled Vehicle} \label{colavoid-pmcv}

Assuming that the controlled vehicle is point-mass, the set $S(\boldsymbol{r}_I^s, \boldsymbol{C}_{BI})$ will reduce to a singleton that can be represented by $\boldsymbol{r}_I^s$ as its position vector, which is bounded by the system's dynamics. Therefore, the distance between two sets, defined by the convex optimization in Eq. (\ref{dist-opt}), will reduce to finding the distance between a point and a set.

\emph{Proposition 2:} Assuming that the controlled vehicle is a point-mass, whose position vector is given by $\boldsymbol{r}_I^s$, and an obstacle $O$ is modeled by either Eqs. (\ref{setC}), (\ref{setD-ex}), or (\ref{setD-app}), the distance between point $s$ and set $O$ is obtained as

\begin{eqnarray} \label{prop2} && \mathrm{min} \; \lVert \boldsymbol{r}_I^s - \boldsymbol{p}_I \rVert^2, \; \mbox{subject to: } f_O(\boldsymbol{p}) \leq 0  \iff \notag \\ && \exists \lambda \geq 0, \mbox{ s.t. } f_O(\tilde{\boldsymbol{p}}) \leq 0, \; \lambda f_O(\tilde{\boldsymbol{p}}) = 0, \; \nabla (\lVert \boldsymbol{r}_I^s - \tilde{\boldsymbol{p}}_I \rVert^2) + \nabla \lambda f_O(\tilde{\boldsymbol{p}}) = 0  \notag \\ \end{eqnarray}

\noindent where $\lambda$ is the optimal dual solution, and $\tilde{\boldsymbol{p}}$ is the optimal primal solution.

\emph{Proof:} Since set $O$ is defined by a differentiable, convex function, and the objective function is convex as well \cite{Boyd2004convex}, the optimization will be categorized as a convex optimization \cite[p. 137]{Boyd2004convex}. Also, set $O$ has a non-empty interior; therefore, the strong duality holds \cite[p. 226]{Boyd2004convex}. As a result, a $\lambda \in \mathbb{R}$, in the case Eq. (\ref{setD-app}) is used, or a $\lambda \in \mathbb{R}^m$, in the case Eq. (\ref{setD-ex}) is used, exists that satisfies the KKT conditions \cite[p. 243]{Boyd2004convex}. Additionally, since the optimization is convex, the KKT conditions satisfy both the necessary and sufficient conditions for optimality \cite[p. 244]{Boyd2004convex}. \hfill $\square$

We can now reformulate the optimal control problem stated in Eq. (\ref{ocp}) using the above results. Hence, the optimal control problem for a point-mass controlled vehicle with the non-conservative, differentiable collision avoidance constraint is given by

\begin{subequations} \begin{eqnarray}  \underset{\boldsymbol{x}, \boldsymbol{u}, t, \tilde{\boldsymbol{p}}^i, \lambda_{i}}{\mathrm{min}} && J = \int_{t_0}^{t_f} {h(\boldsymbol{x}, \boldsymbol{u}, t) \; dt} \\ \mbox{subject to:} && \boldsymbol{x}_0 = \boldsymbol{x}_I, \; \boldsymbol{x}_f = \boldsymbol{x}_T \\ && \dot{\boldsymbol{x}} = \boldsymbol{f}(\boldsymbol{x}, \boldsymbol{u}, t) \\ && \boldsymbol{g}(\boldsymbol{x}, \boldsymbol{u}, t) \leq 0 \\ && \lambda_{i} \geq 0, \; f_{O_i}(\tilde{\boldsymbol{p}}^i) \leq 0, \; \lambda_{i} f_{O_i}(\tilde{\boldsymbol{p}}^i) = 0 \\ && \nabla (\lVert \boldsymbol{r}_I^s - \tilde{\boldsymbol{p}}^i_I \rVert^2) + \nabla \lambda_i f_{O_i}(\tilde{\boldsymbol{p}}^i) = 0 \\ &&  \lVert \boldsymbol{r}_I^s - \tilde{\boldsymbol{p}}^i_I \rVert^2 \geq d_{safe}^2, \; i = 1, ..., n_o \end{eqnarray} \label{pm-ocp} \end{subequations}

\noindent where $d_{safe}$ is the minimum safety distance set by the designer, depending on the application, and $n_o$ is the number of obstacles. Note that the optimization now takes place over $\tilde{\boldsymbol{p}}^i$, the corresponding closest point of set $O_i$ to the controlled vehicle, and $\lambda_{i}$, the dual variable, along with the state $\boldsymbol{x}$ and control input $\boldsymbol{u}$. According to the definition of set $O_i$, the dual variable is always one-dimensional when Eq. (\ref{setD-app}) is used and m-dimensional when Eq. (\ref{setD-ex}) is used; however, the dimension of $\tilde{\boldsymbol{p}}^i$ depends on the vector space dimension, where the optimization is performed. An advantage of such a formulation is that the distance between the controlled vehicle and the obstacle can be established at each time instant, no matter how close they are.\\


\subsection{Collision Avoidance For Full-dimensional Controlled Vehicle} \label{colavoid-fdcv}

Assuming that the controlled vehicle has a full-dimensional shape, it, at least, consists of one convex, compact set. The position and orientation of set $S$ are determined by $\boldsymbol{r}_I^s$ and $\boldsymbol{C}_{BI}$, respectively, which are bounded by the system's dynamics.

\emph{Proposition 3:} Assuming a full-dimensional controlled vehicle $S(\boldsymbol{r}_I^s, \boldsymbol{C}_{BI})$, whose position and orientation are given by $\boldsymbol{r}_I^s$ and $\boldsymbol{C}_{BI}$, respectively, along with an obstacle $O$ are modeled by either Eqs. (\ref{setC}), (\ref{setD-ex}), or (\ref{setD-app}), the distance between sets $S(\boldsymbol{r}_I^s, \boldsymbol{C}_{BI})$ and $O$ is obtained as

\begin{eqnarray} \label{prop3} && \mathrm{min} \; \lVert \boldsymbol{w}_I - \boldsymbol{p}_I \rVert^2, \; \mbox{subject to: } f_S(\boldsymbol{w}) \leq 0, \; f_O(\boldsymbol{p}) \leq 0   \iff  \notag \\  && \exists \lambda, \mu \geq 0, \; \mathrm{s.t.} \; f_S(\tilde{\boldsymbol{w}}) \leq 0, \; f_O(\tilde{\boldsymbol{p}}) \leq 0, \; \lambda f_S(\tilde{\boldsymbol{w}}) = 0, \; \mu f_O(\tilde{\boldsymbol{p}}) = 0 \notag \\ && \nabla (\lVert \tilde{\boldsymbol{w}}_I - \tilde{\boldsymbol{p}}_I \rVert^2) + \nabla \lambda f_S(\tilde{\boldsymbol{w}}) + \nabla \mu f_O(\tilde{\boldsymbol{p}}) = 0 \end{eqnarray}
\vspace{0.5em}

\noindent where $\lambda$ and $\mu$ are dual optimal solution, and $\tilde{\boldsymbol{w}}$ and $\tilde{\boldsymbol{p}}$ are primal optimal solution.

\emph{Proof:} The optimization given in the proposition statement is convex since the objective function is convex, as well as the inequality constraints, as sets $S$ and $O$ are both defined by differentiable convex functions \cite[p. 137]{Boyd2004convex}. Additionally, sets $S$ and $O$ have non-empty interiors, so the strong duality holds \cite[p. 226]{Boyd2004convex}. Therefore, there exist $\lambda, \mu \in \mathbb{R}$, in the case Eq. (\ref{setD-app}) is used, or $\boldsymbol{\lambda}, \boldsymbol{\mu} \in \mathbb{R}^m$, in the case Eq. (\ref{setD-ex}) is used, that satisfy the KKT conditions \cite[p. 243]{Boyd2004convex}, which are the necessary and sufficient conditions for optimality due to the convexity of the optimization \cite[p. 244]{Boyd2004convex}. \hfill $\square$

Having obtained the new set of differentiable constraints for collision avoidance for full-dimensional controlled vehicles, the optimal control problem, including the reformulated constraints, is given by

\begin{subequations} \begin{eqnarray}  \underset{\boldsymbol{x}, \boldsymbol{u}, t, \tilde{\boldsymbol{w}}^{ij}, \tilde{\boldsymbol{p}}^{ij}, \lambda_{ij}, \mu_{ij}}{\mathrm{min}} && J = \int_{t_0}^{t_f} {h(\boldsymbol{x}, \boldsymbol{u}, t)\; dt} \\ \mbox{subject to:} && \boldsymbol{x}_0 = \boldsymbol{x}_I, \; \boldsymbol{x}_f = \boldsymbol{x}_T \\  && \dot{\boldsymbol{x}} = \boldsymbol{f}(\boldsymbol{x}, \boldsymbol{u}, t) \\ && \boldsymbol{g}(\boldsymbol{x}, \boldsymbol{u}, t) \leq 0 \\ && \lambda_{ij} \geq 0, \; f_{S_i}(\tilde{\boldsymbol{w}}^{ij}) \leq 0, \; \lambda_{ij} f_{S_i}(\tilde{\boldsymbol{w}}^{ij}) = 0  \\ && \mu_{ij} \geq 0, \; f_{O_j}(\tilde{\boldsymbol{p}}^{ij}) \leq 0, \; \mu_{ij} f_{O_j}(\tilde{\boldsymbol{p}}^{ij}) = 0 \\ && \nabla (\lVert \tilde{\boldsymbol{w}}^{ij}_I - \tilde{\boldsymbol{p}}^{ij}_I \rVert^2) + \nabla \lambda_{ij} f_{S_i}(\tilde{\boldsymbol{w}}^{ij}) + \nabla \mu_{ij} f_{O_j}(\tilde{\boldsymbol{p}}^{ij}) = 0 \label{diff-cond-ocp}  \\ && \lVert \tilde{\boldsymbol{w}}^{ij}_I - \tilde{\boldsymbol{p}}^{ij}_I \rVert^2 \geq d_{safe}^2, \; i = 1,...,n_s, \; j = 1,..., n_o \label{d-safe-ocp} \end{eqnarray} \label{fd-ocp} \end{subequations}

\noindent where $n_s$ and $n_o$ are the number of robotic spacecraft components and obstacles, respectively, and $d_{safe}$ is the minimum safety distance set by the designer.  $\lambda_{ij}, \mu_{ij} \in \mathbb{R}$ are the dual variables introduced for each component pair. Also, $\tilde{\boldsymbol{w}}^{ij}$ is another optimization parameter added to the problem, which is the corresponding closest point of set $S_i$ w.r.t. set $O_j$, as is vice versa for point $\tilde{\boldsymbol{p}}^{ij}$. Therefore, the optimization is now performed over the optimal dual variables ($\lambda_{ij}$, $\mu_{ij}$), the closest points on the controlled vehicle parts and obstacles ($\tilde{\boldsymbol{w}}^{ij}$, $\tilde{\boldsymbol{p}}^{ij}$) along with the state $\boldsymbol{x}$ and control input $\boldsymbol{u}$. According to the definition of sets $S$ and $O_i$, the dual variables are always one-dimensional when Eq. (\ref{setD-app}) is used and m-dimensional when Eq. (\ref{setD-ex}) is used; however, the dimensions of $\tilde{\boldsymbol{w}}^i$ and  $\tilde{\boldsymbol{p}}^i$ depend on the vector space dimension, where the optimization is performed on. \\


\section{Autonomous On-Orbit Assembly of Complex Space Structures} \label{app-aoa}

As mentioned earlier, the proposed technique above was initially inspired by how to perform an autonomous on-orbit assembly with a non-conservative collision avoidance constraint. Hence, we apply the proposed technique for two assembly scenarios in tight environments involving a complex structure with multiple components. Since collision avoidance in an AOA operation is a serious constraint, the robotic spacecraft and the target structure must be modeled as accurately as possible, and the resulting constraint should be the least conservative so that the optimal control problem would be able to find a feasible trajectory. Intuitively, it is apparent that conservative models, such as ellipsoids, would lead to roadblocks in finding a feasible trajectory in tight environments.


\subsection{Dynamics, Objective, Environment}

This section will derive the equations of motion of the robotic spacecraft maneuvering close to the target space structure. Without loss of generality, it is assumed that the operation takes place in deep space where the spacecraft is only under the influence of the internal forces produced by thrusters, reaction wheels, etc. Due to the micro-gravitational effects of the corresponding environment, the orbital dynamics can be suitably ignored. As a result, the spacecraft's representative state variables are force-independent, and the relative dynamics can be reduced to double integrator \cite{scharf2004}. Therefore, for a given robotic spacecraft shown in Fig. \ref{robotic-sc}, we can resolve the position and velocity of the spacecraft in the inertial frame $\boldsymbol{\mathcal{F}}_I$ as

\begin{equation} \label{pos-vel-sc} \boldsymbol{r}^s = \boldsymbol{\mathcal{F}}_I^T \boldsymbol{r}_I^s, \quad \dot{\boldsymbol{r}}^s = \boldsymbol{\mathcal{F}}_I^T \dot{\boldsymbol{r}}_I^s \end{equation}


\tikzset{
	manipulator/.pic ={
		\path[shape=coordinate]
			(0,0) coordinate(o)
			[rotate=30] (15mm, 5mm) coordinate(size)
			[rotate=0] (15mm, 2.5mm) coordinate(a)
			[rotate=0] (25mm, 12.5mm) coordinate(b)
			(b) +(0mm, -2.5mm) coordinate(c);

		\draw[rotate=30] (o) rectangle (size);
		\draw[dashed, thick, anchor=mid, rotate=0] (a) -- (b); 
		\draw[rotate=30] (c) rectangle +(15mm, 5mm); 
	}
}

\begin{figure}[h]
\centering
\begin{tikzpicture}
\draw (-4, -2) .. controls (-4,-1) and (4,-1) .. (4, -3) pic[midway, scale=0.6] {manipulator}
			.. controls (4,-4) and (-4,-3) .. 
			pic[at start, xshift=3pt, scale=0.6, right] {manipulator} 
			pic[midway, yshift=-3mm, scale=0.6, rotate=-60] {manipulator} 
			node[midway, xshift=4pt] (contact p) {}
			node[midway, yshift = -12mm, xshift = 15mm, font=\footnotesize] {$i^{th}$ Robotic Manipulator}
			cycle;

\filldraw[black] (contact p) circle (2pt); 

\draw[thick, ->] (-6, -3) -- (-6, -2);
\draw[thick, ->] (-6, -3) -- (-5, -3);
\draw[thick, ->] (-6, -3) -- (-6.7, -3.7);
\node[anchor=north west] at (-6,-3) {$\boldsymbol{\mathcal{F}}_I$};

\fill (-3, -2) node (s) {} circle (0.05cm) node[anchor=north] {$s$};
\draw[thick, ->] (-6, -3) -- (-3,-2);

\draw[thick, ->] (-3,-2) -- (-2.3, -1.3);
\draw[thick, ->] (-3,-2) -- (-3.7, -1.3);
\draw[thick, ->] (-3,-2) -- (-3, -1) node[anchor=east] {$\boldsymbol{\mathcal{F}}_s$};

\draw[thick, ->] (s.center) -- (contact p.center) node[very near end, above] {$\boldsymbol{\rho^{m_i}_s}$};
\draw[thick, <-] (contact p.center) -- +(-7.5mm, -7.5mm) node[near end, above] {$\boldsymbol{g}_s^{m_i}$};
\draw[thick, <-] (contact p.center) -- +(9.5mm, 2mm) node[midway, above] {$\boldsymbol{f}_s^{m_i}$};
\end{tikzpicture}
\caption{Robotic Spacecraft}
\label{robotic-sc}
\end{figure}
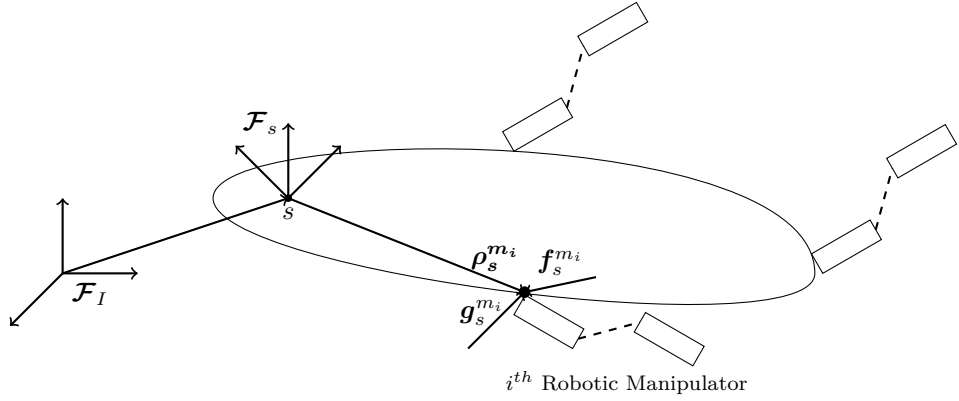

The total kinetic energy of the robotic spacecraft can be written as

\begin{equation} T = \frac{1}{2} m_t {\boldsymbol{v}_s^s}^T \boldsymbol{v}_s^s + \frac{1}{2} \boldsymbol{J}_{t,s} {\boldsymbol{\omega}_s^s}^T \boldsymbol{\omega}_s^s - {\boldsymbol{v}_s^s}^T \boldsymbol{c}_{t,s}^{\times} \boldsymbol{\omega}_s^s \end{equation}

\noindent where $m_t$ is the total mass of the spacecraft, $\boldsymbol{J}_{t,s}$ is the total moment of inertia matrix about point $s$ evaluated in $\boldsymbol{\mathcal{F}}_s$, and $\boldsymbol{c}_{t,s}$ is the total first moment of mass about point $s$ evaluated in $\boldsymbol{\mathcal{F}}_s$. In addition, $\boldsymbol{v}_s^s = \boldsymbol{C}_{sI}^T \boldsymbol{r}_I^s$ and $\boldsymbol{\omega}_s^s$ are the inertial and angular velocity of the spacecraft resolved in $\boldsymbol{\mathcal{F}}_s$, respectively, where $\boldsymbol{C}_{sI}$ is the rotation matrix transforming coordinates from the inertial frame to the body-fixed frame.

Since there is no potential energy due to being in the deep space environment, the equations of motion are given using the Lagrange method as

\begin{eqnarray} \label{lagrange-eqs-m} && \frac{d}{dt}(\frac{\partial{T}}{\partial{\boldsymbol{v}_s^s}}) + {\boldsymbol{\omega}_s^s}^{\times} \frac{\partial{T}}{\partial{\boldsymbol{v}_s^s}} = \boldsymbol{f}_s + \sum_{i=1}^{n} {\boldsymbol{f}_s^{m_i}} \notag \\ && \frac{d}{dt}(\frac{\partial{T}}{\partial{\boldsymbol{\omega}_s^s}}) + {\boldsymbol{v}_s^s}^{\times} \frac{\partial{T}}{\partial{\boldsymbol{v}_s^s}} + {\boldsymbol{\omega}_s^s}^{\times} \frac{\partial{T}}{\partial{\boldsymbol{\omega}_s^s}} = \boldsymbol{g}_s + \sum_{i=1}^{n} {\boldsymbol{g}_s^{m_i} + \boldsymbol{\rho}_s^{m_i \times} \boldsymbol{f}_s^{m_i}} \end{eqnarray}

\noindent where $\boldsymbol{f}_s$ and $\boldsymbol{g}_s$ are the total thruster forces and reaction wheel torques exerted on the spacecraft, respectively, while $\boldsymbol{f}_s^{m_i}$ and $\boldsymbol{g}_s^{m_i}$ are the forces and torques applied to the body of the robotic spacecraft by the attached manipulators, respectively. Working the equations of motion given in Eq. (\ref{lagrange-eqs-m}) results in

\begin{equation} \boldsymbol{M}_t \dot{\boldsymbol{\nu}}_s^s + \boldsymbol{l}(\boldsymbol{\nu}_s^s) = \boldsymbol{q}_s + \boldsymbol{q}_s^m \end{equation}

\noindent where

\begin{equation} \boldsymbol{\nu}_s^s = \begin{bmatrix} \boldsymbol{v}_s^s \\ \boldsymbol{\omega}_s^s \end{bmatrix}, \quad \boldsymbol{M}_t = \begin{bmatrix} m_t \boldsymbol{I} & -\boldsymbol{c}_{t,s}^{\times} \\ \boldsymbol{c}_{t,s}^{\times} & \boldsymbol{J}_{t,s} \end{bmatrix}, \quad \boldsymbol{l}(\boldsymbol{\nu}_s^s) = \begin{bmatrix} m_t {\boldsymbol{\omega}_s^s}^{\times} \boldsymbol{v}_s^s \\ ({\boldsymbol{v}_s^s}^{\times}{\boldsymbol{\omega}_s^s}^{\times} - {\boldsymbol{\omega}_s^s}^{\times} {\boldsymbol{v}_s^s}^{\times}) \boldsymbol{c}_{t,s} + {\boldsymbol{\omega}_s^s}^{\times} \boldsymbol{J}_{t,s} \boldsymbol{\omega}_s^s \end{bmatrix}, \notag \end{equation}

\begin{equation} \boldsymbol{q}_s = \begin{bmatrix} \boldsymbol{f}_s \\ \boldsymbol{g}_s \end{bmatrix}, \quad \boldsymbol{q}_s^m = \begin{bmatrix} \sum_{i=1}^{n} {\boldsymbol{f}_s^{m_i}} \\ \sum_{i=1}^{n} {\boldsymbol{g}_s^{m_i} + \boldsymbol{\rho}_s^{m_i \times} \boldsymbol{f}_s^{m_i}} \end{bmatrix} \notag \end{equation}

The associated kinematics using the unit quaternion parametrization $(\eta, \boldsymbol{\epsilon})$ for the attitude of the spacecraft are

\begin{equation} \dot{\boldsymbol{r}}_I^s = \boldsymbol{C}_{sI}^T \boldsymbol{v}_s^s, \quad \dot{\boldsymbol{\epsilon}} = \frac{1}{2} (\eta \boldsymbol{I} - \boldsymbol{\epsilon}^{\times}) \boldsymbol{\omega}_s^s, \quad \dot{\eta} = -\frac{1}{2} \boldsymbol{\epsilon}^T \boldsymbol{\omega}_s^s \end{equation}

\noindent where

\begin{equation} \boldsymbol{C}_{sI} = (\eta^2 - \boldsymbol{\epsilon}^T \boldsymbol{\epsilon}) \boldsymbol{I} + 2 \boldsymbol{\epsilon} \boldsymbol{\epsilon}^T - 2 \eta \boldsymbol{\epsilon}^{\times} \end{equation}

We further consider a robotic spacecraft with no manipulator in our numerical case studies in order to omit unnecessary complications of the AOA missions from the problem to better capture the capabilities and potential of the current work. Therefore, the only modification to our equations of motion would be the removal of the contact forces and torques of the manipulators on the spacecraft's body $\boldsymbol{q}_s^m$. For proof-of-concept purposes, the proposed method is evaluated both in proximity operations in general and, more specifically, in the AOA operations using two scenarios in our numerical example. Both scenarios aim to navigate the robotic spacecraft in a tight environment to the assembly point in close proximity to the target structure with the least fuel consumption possible. Consequently, the performance index can be written as

\begin{equation} \label{objective-u}  J = \int_{t_0}^{t_f} {\lVert \boldsymbol{u}(t) \rVert_2^2 \; dt} \end{equation}

\noindent where $\boldsymbol{u}(t) = \begin{bmatrix} f_{x,s} & f_{y,s} & f_{z,s} & g_{x,s} & g_{y,s} & g{z,s} \end{bmatrix}^T$. Moreover, the available thrust is bounded by the maximum available thrust $f_{max}$, while the available torque is capped by the maximum available torque $g_{max}$; therefore, the corresponding constraints are defined as

\begin{equation} f_{i,s}(t)^2 \leq f_{max}^2, \quad g_{i,s}(t)^2 \leq g_{max}^2, \quad i = \{x, y, z\} \end{equation}

For differentiability purposes, the Euclidean norm is utilized for defining the distance between any two objects; therefore, the minimum safety distance constraint in Eq. (\ref{d-safe-ocp}) is written as

\begin{equation} \lVert \tilde{\boldsymbol{w}}^{ij}_I - \tilde{\boldsymbol{p}}_I^{ij} \rVert_2^2  \geq  d_{safe}^2 \end{equation}


\subsection{Robotic Spacecraft and Target structure Models} \label{sc-st-models}

\subsubsection*{Scenario A}

The robotic spacecraft unit is a $4 m \times 2 m \times 2 m$ cuboid in the first scenario. On the other hand, the target structure consists of two components, a rectangular cuboid with a size of 2 m, 6 m, and 6 m and a cylinder with a base diameter of 2 m and height of 4 m. The spacecraft and structure are modeled with respect to their body-fixed frames using Eqs. (\ref{c-func}) and (\ref{d-func3}) as

\begin{eqnarray}  && f_S(\boldsymbol{w}) = ln[(e^{(x_w-2)8} + e^{(y_w-1)8} + e^{(z_w - 1)8} + e^{(-x_w-2)8} + e^{(-y_w-1)8} + e^{(-z_w - 1)8})^{\frac{1}{8}}]  \notag \\ && f_{T_1}(\boldsymbol{p}^1) = ln[(e^{(x_{T_1}-1)8} + e^{(y_{T_1}-3)8} + e^{(z_{T_1} - 3)8} + e^{(-x_{T_1}-2)8} + e^{(-y_{T_1}-3)8} + e^{(-z_{T_1} - 3)8})^{\frac{1}{8}}]  \notag \\ && f_{T_2}(\boldsymbol{p}^2) = (\frac{x_{T_2}}{2})^8 + (y_{T_2})^2 + (z_{T_2})^2 - 1  \end{eqnarray} \label{scenA-models}

\noindent where the target structure components $T_1$ and $T_2$ are located at $\boldsymbol{d}_I^{T_1}=(5, 0,0)$m and $\boldsymbol{d}_I^{T_2}=(8,0,0)$m. Although the target structure can have a rotational and translational motion, it is assumed that it is stationary for both scenarios. However, the inertial rotation and translation of the moving robotic spacecraft can be resolved as

\begin{equation}\label{inertial-sc} \boldsymbol{w}_I = \boldsymbol{C}_{sI}^T \boldsymbol{w} + \boldsymbol{r}_I^s \end{equation}

On the other hand, since the body-fixed frames for the target structure components are assumed to be aligned with the inertial frame; therefore, $\boldsymbol{C}_{IB}$ is equal to identity, and the inertial coordinates can be obtained as

\begin{equation}\label{inertial-st} \boldsymbol{\mathcal{F}}_I^T \boldsymbol{p}^i_I = \boldsymbol{\mathcal{F}}_I^T \boldsymbol{C}_{IB}^i \boldsymbol{p}_B^i + \boldsymbol{\mathcal{F}}_I^T \boldsymbol{d}_{I}^{i}, \quad i = \{ T_1,T_2 \} \end{equation}
\vspace{0.5em}

As mentioned earlier, using infinity exponents to model the polyhedral shapes is numerically infeasible. Therefore, the exponent of 8 was chosen for modeling the cuboids in these scenarios. Albeit larger exponents can be selected depending on the mission accuracy requirements. The exponents only affect the corners; consequently, the larger the exponents, the sharper the corners would be. This approximation can be compensated by choosing an appropriate $d_{safe}$ at corners. However, as will be discussed in the discussion section, the corners are desirably sharp enough for assembly purposes with the exponent of 8. This is the sacrifice to significantly decrease the optimal control problem's dimension. In the case that the exact modeling of corners is desired, Eq. (\ref{d-func}) may be used. Section \ref{discussion} will present an analysis on the exponent effects on the modeling accuracy of corners.

\subsubsection*{Scenario B}

The identical robotic spacecraft as in scenario A is used for the second numerical case study. The target structure consists of three components, with two of those being rectangular cuboids with a size of (4 m, 2 m, 2 m) and (2 m, 8 m, 2 m), respectively, and the last one being a cylinder with the base diameter of 2 m and height of 4 m. The respective functions defined with respect to the corresponding body-fixed frames are

 \begin{eqnarray} && f_{T_1}(\boldsymbol{p}^1) = ln[(e^{(x_{T_1}-2)8} + e^{(y_{T_1}-1)8} + e^{(z_{T_1} - 1)8} + e^{(-x_{T_1}-2)8} + e^{(-y_{T_1}-1)8} + e^{(-z_{T_1} - 1)8})^{\frac{1}{8}}]  \notag \\ && f_{T_2}(\boldsymbol{p}^2) = ln[(e^{(x_{T_2}-1)8} + e^{(y_{T_2}-4)8} + e^{(z_{T_2} - 1)8} + e^{(-x_{T_2}-1)8} + e^{(-y_{T_2}-4)8} + e^{(-z_{T_2} - 1)8})^{\frac{1}{8}}]  \notag \\ && f_{T_3}(\boldsymbol{p}^3) = (\frac{x_{T_3}}{2})^8 + (y_{T_3})^2 + (z_{T_3})^2 - 1 \end{eqnarray}

\noindent where the structure components are located at $\boldsymbol{d}_I^{T_1} = (10,3, 0)$m, $\boldsymbol{d}_I^{T_2} = (13,0,0)$m, and $\boldsymbol{d}_I^{T_3} = (10,-3,0)$m. The inertial coordinates for the spacecraft and the target structure components are obtained as in Eqs. (\ref{inertial-sc}) and (\ref{inertial-st}), respectively, where $i = \{1,2,3 \}$ in this case.


\subsection{Simulation Results}

To examine the performance of the proposed technique, the simulation results of the two scenarios are presented and discussed in this section. Each scenario was solved using a MATLAB toolbox based on the pseudospectral method for solving optimal control problems \cite{ross2015primer}. It implements a fast spectral algorithm by expanding the infinite series of special basis functions, which is truncated at the desired convergence tolerance \cite{ross2012review}. The optimal control problems were solved on a Dell Inspiron with a 2.8 GHz core-i7 processor.

\subsubsection*{Scenario A}

\begin{figure}[H]
\centering
	\begin{subfigure}{0.5\textwidth}
	\includegraphics[width =0.9 \linewidth, height = 6cm]{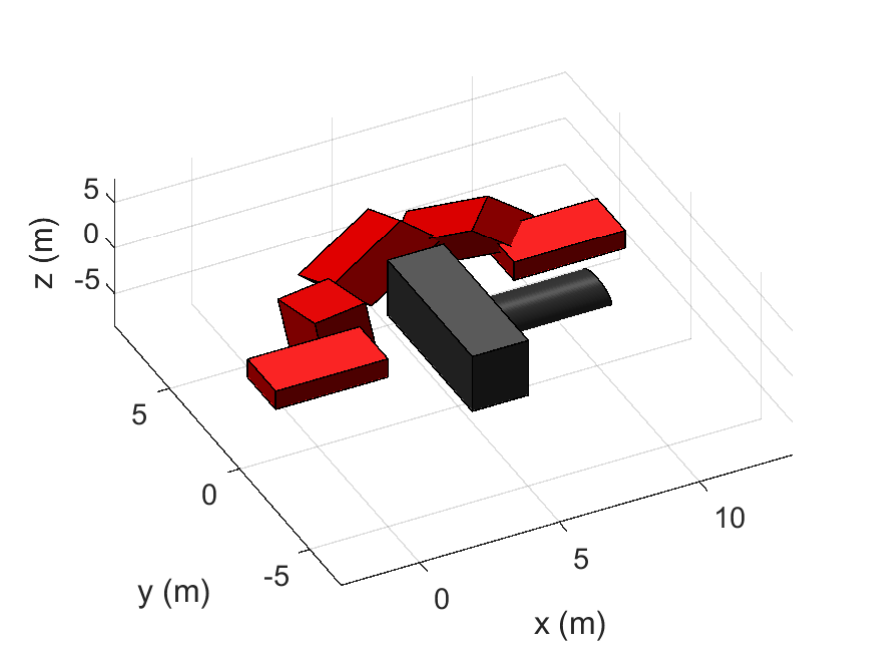}
	\caption{ View A}
	\end{subfigure}%
	\begin{subfigure}{0.5\textwidth}
	\includegraphics[width =0.9 \linewidth, height = 6cm]{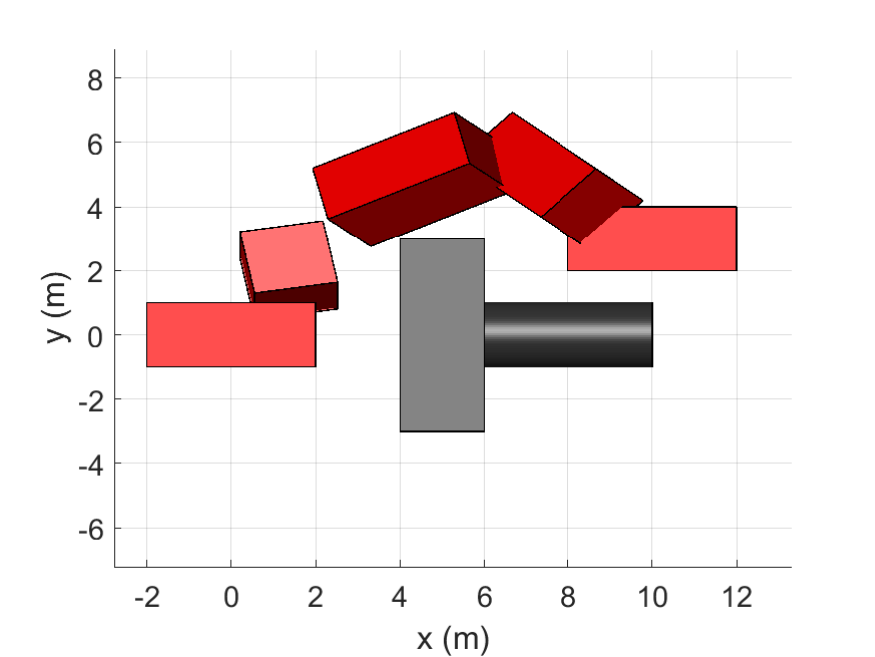}
	\caption{View B}
	\end{subfigure}
\caption{Robotic Spacecraft Trajectory In Close Proximity of Target Structure - Scenario A: Red Object (Spacecraft) - Black Objects (Target Structure)}
\label{trajA}
\end{figure}

 In Scenario A, the target structure's setup of its components is shown in \figurename \ref{trajA}. The robotic spacecraft should start from a stationary initial condition $(0, 0, 0)$ m and arrive at rest at the desired assembly point $(10, 3, 0)$ m. The safety distance $d_{safe}$ for this scenario is set to 0.1 m, which is a non-conservative value for such an operation. Based on the size of the spacecraft and target structure, the defined $d_{safe}$, and the terminal conditions, the distance between the spacecraft and the structure becomes the minimum possible at many points within the trajectory. The operation time was set to 150 s, and the maximum available thrust and torque were selected as 0.02 N and 0.01 N.m, respectively. Also, the robotic spacecraft unit's mass and inertia were considered 3 kg and 5 kg/m$^2$, respectively. Additionally, 20 nodes were used for discretization to run Scenario A.

\begin{figure}[h!]
\centering
	\begin{subfigure}{\textwidth}
	\centering
	\includegraphics[width=\textwidth, height=10cm]{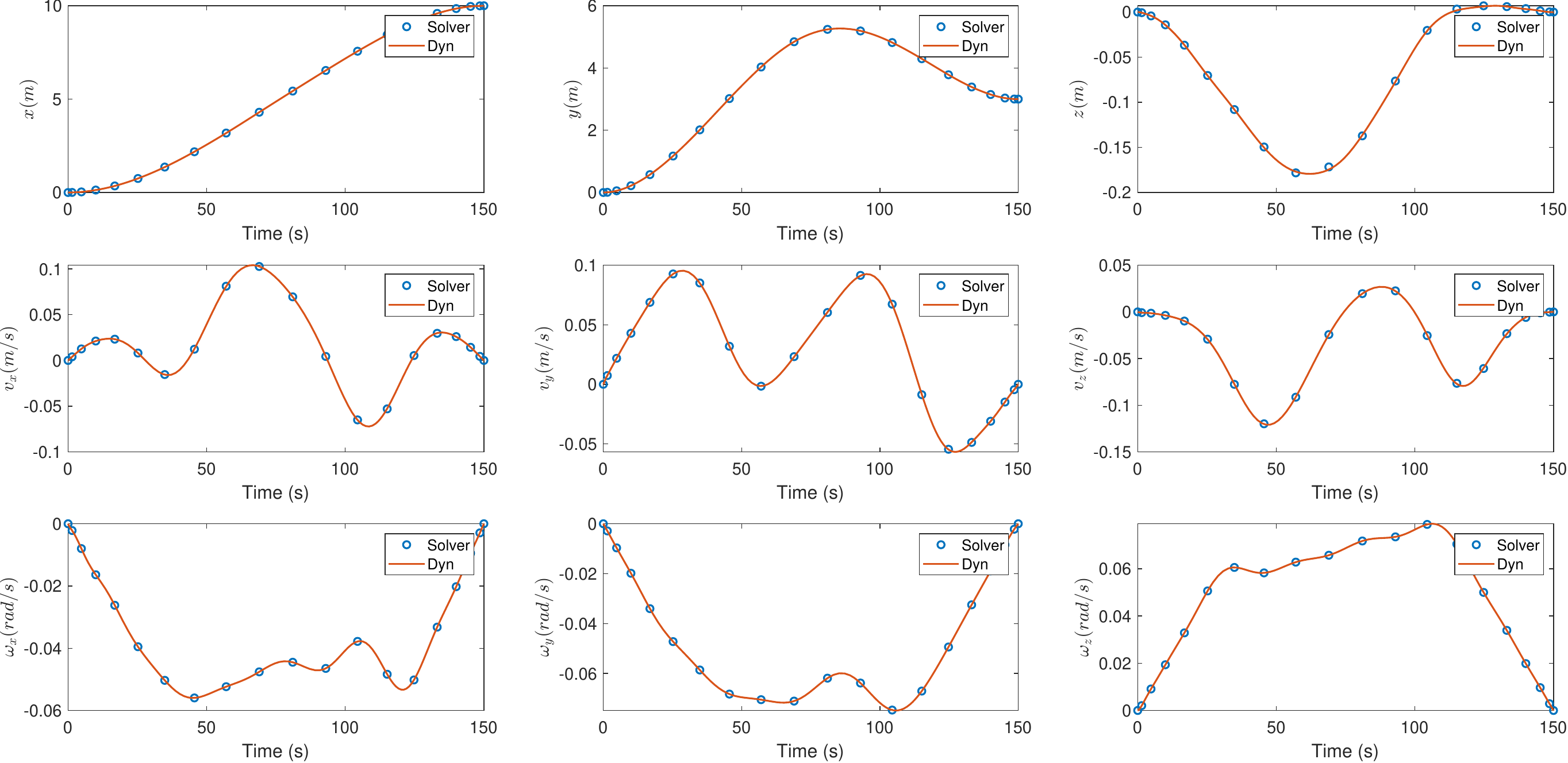}
	\caption{ States: Position (Row 1), Translational Velocity (Row 2) and Angular Velocity (Row 3) }
	\end{subfigure}
	\newline
	\begin{subfigure}{\textwidth}
	\includegraphics[width=\textwidth, height = 7cm]{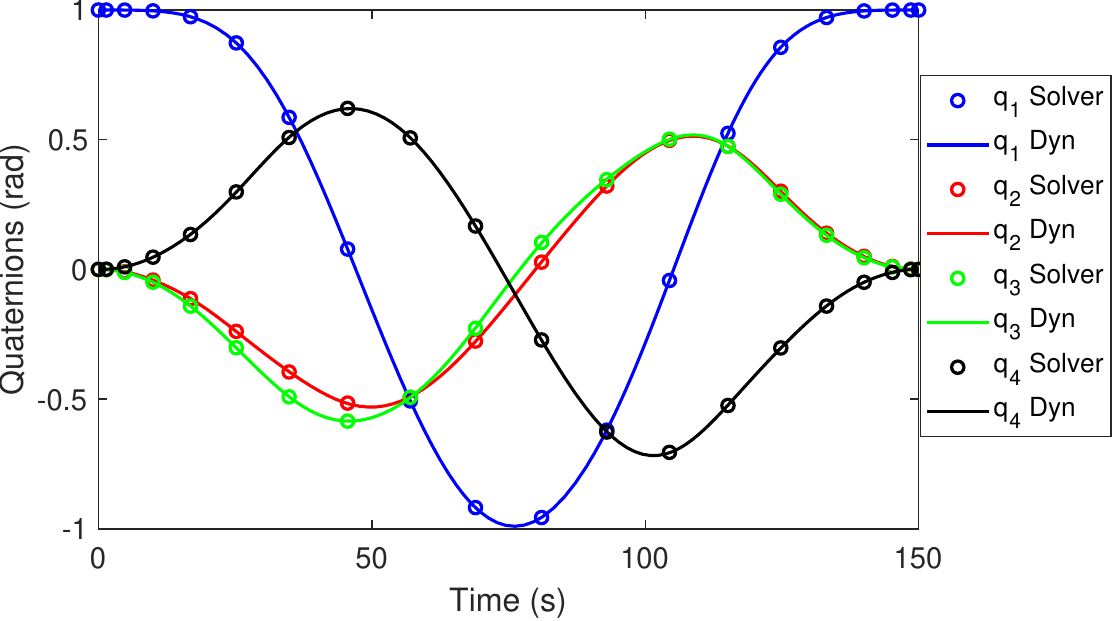}
	\caption{Unit Quaternions}
	\end{subfigure}
\caption{States Time History - Scenario A}
\label{scenA-state}
\end{figure}

\begin{figure}[h]
\centering
\includegraphics[width =  \textwidth, height = 10cm]{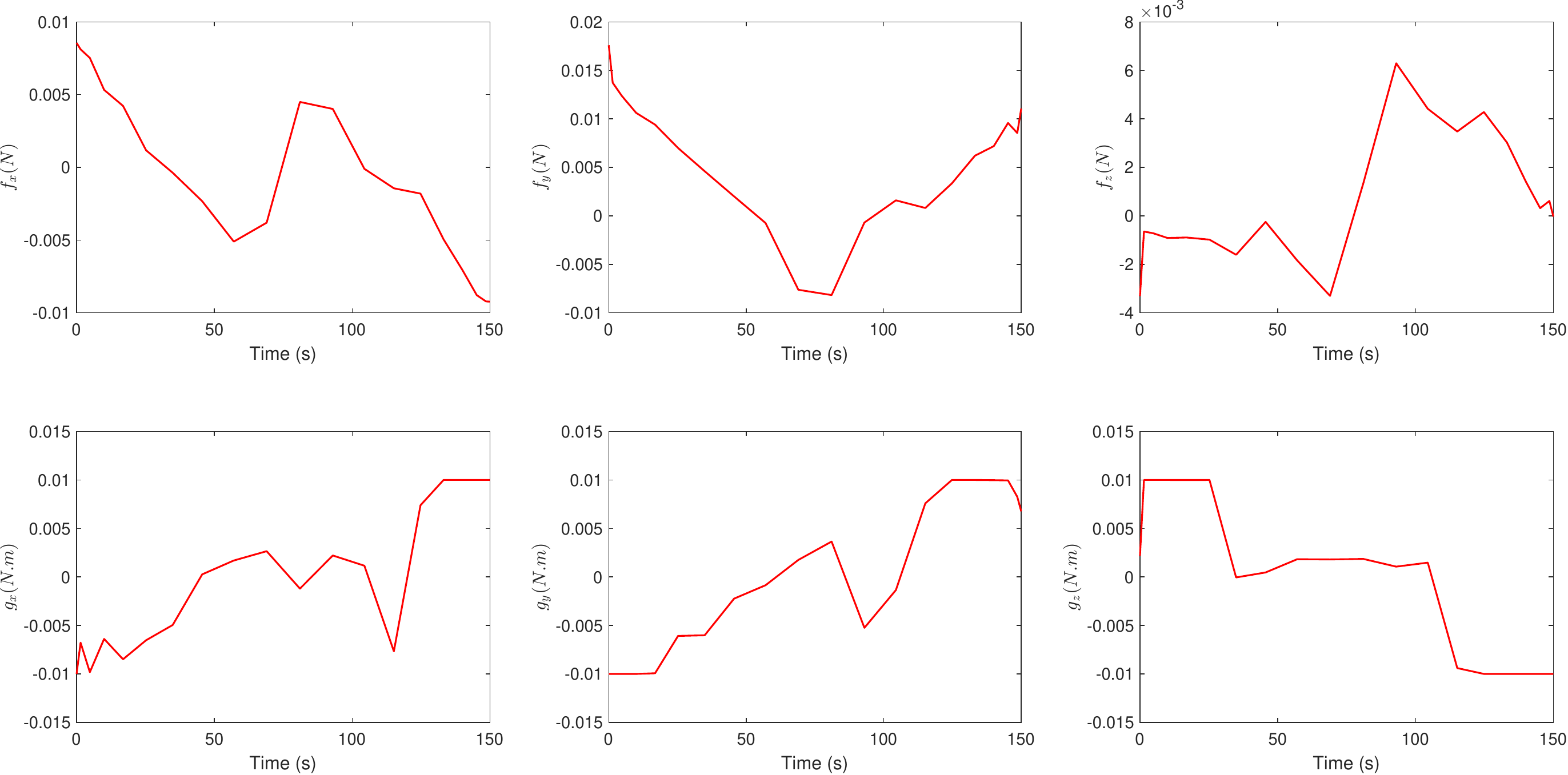}
\caption{Control Inputs Time History - Scenario A}
\label{scenA-controls}
\end{figure}

The results in \figurename \ref{trajA} illustrate the effectiveness and accuracy of the formulation and the fact that the desired optimal trajectory can be found for such a close-proximity mission. It also demonstrates how closely the spacecraft maintains its distance with respect to the target structure while maneuvering around it without violating the collision avoidance constraint $d \geq d_{safe}$. The time history of the state and control input are illustrated in \figurename \ref{scenA-state} and \figurename \ref{scenA-controls}. We see that the state exactly meets the terminal conditions, although the assembly point brings the spacecraft to the boundary of the allowed distance with the structure. Additionally, the control input time history shows that the controls stuck to the allowed region limited by the maximum available thrust and torque.

\begin{figure}[H]
\centering
	\begin{subfigure}{0.5\textwidth}
	\includegraphics[width =0.9 \linewidth, height = 6cm]{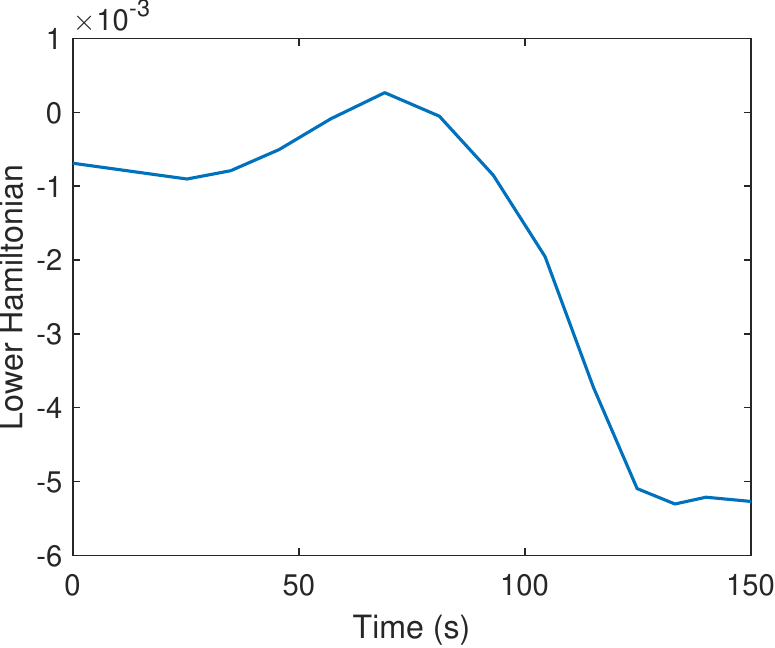}
	\caption{ Scenario A}
	\label{lowhamA}
	\end{subfigure}%
	\begin{subfigure}{0.5\textwidth}
	\includegraphics[width =0.9 \linewidth, height = 6cm]{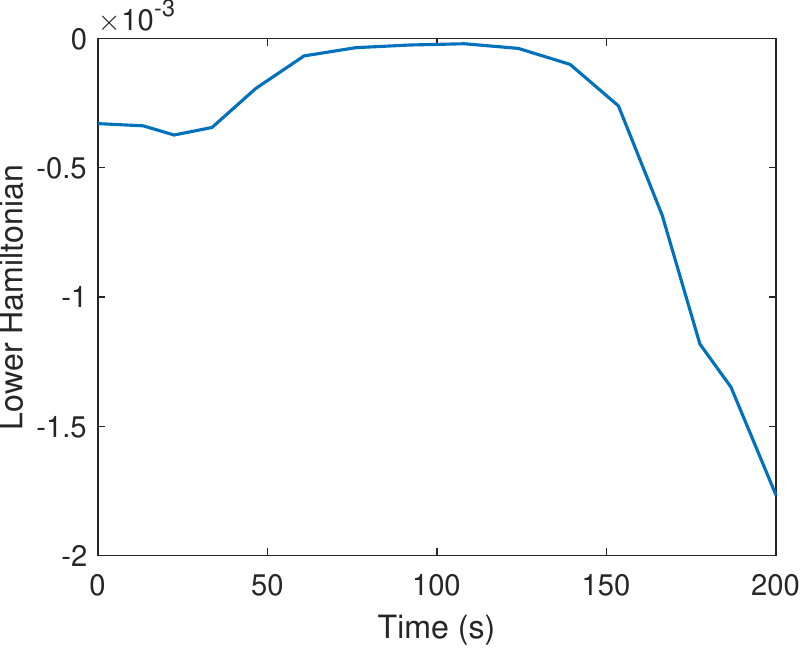}
	\caption{Scenario B}
	\label{lowhamB}
	\end{subfigure}
\caption{Lower Hamiltonian Time History}
\label{lowham}
\end{figure}

Although the necessary conditions for optimality could be developed to validate the results against them, the variation of the corresponding lower Hamiltonian is demonstrated in \figurename \ref{lowhamA}. It can be seen that the lower Hamiltonian is constant, and the obtained solution is indeed a Pontryagin's extremal.   

It is also a common practice to validate the solver's output by integrating the system's dynamics using the output. The results of such a practice can be seen in \figurename \ref{scenA-state}a, where the 'o' marks represent the solver solutions and the continuous graphs show the dynamics results. The figure demonstrates that both results are consistent, and the solver's solution can be accepted as a feasible optimal solution.

\subsubsection*{Scenario B}

\begin{figure}[h]
\centering
	\begin{subfigure}{0.5\textwidth}
	\includegraphics[width =0.9 \linewidth, height = 6cm]{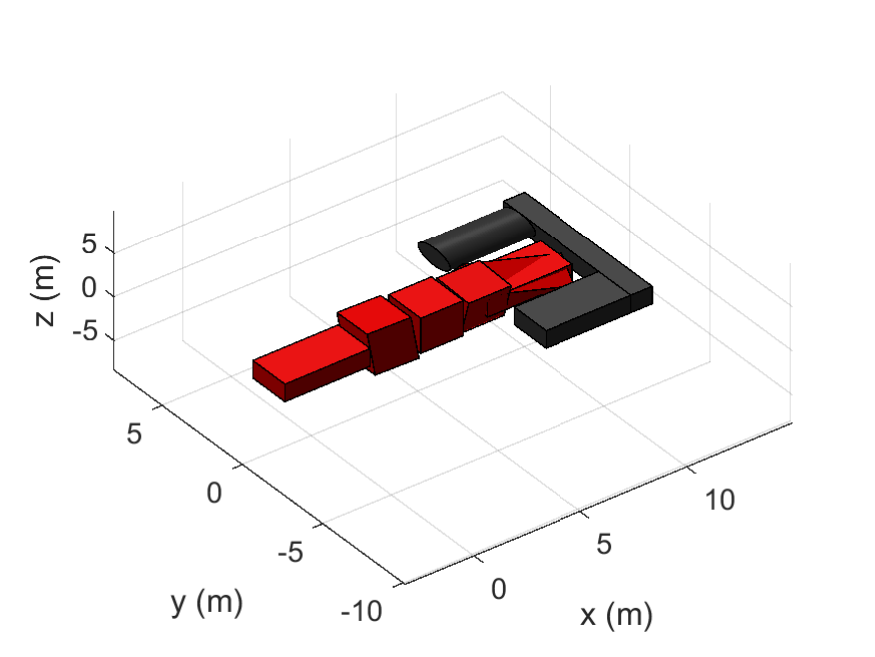}
	\caption{ View A}
	\end{subfigure}%
	\begin{subfigure}{0.5\textwidth}
	\includegraphics[width =0.9 \linewidth, height = 6cm]{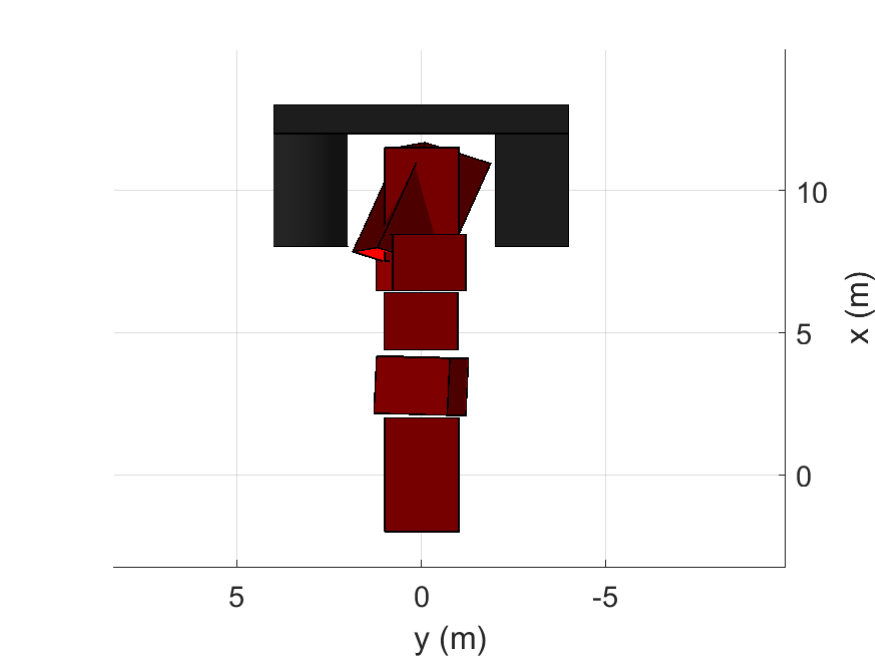}
	\caption{View B}
	\end{subfigure}
\caption{Robotic Spacecraft Trajectory in AOA mission with Target Structure - Scenario B: Red Object (Spacecraft) - Black Objects (Target Structure)}
\label{trajB}
\end{figure}

Similar to Scenario A, in this section, the robotic spacecraft unit has the same properties, and the target structure is built as presented in \figurename\ref{trajB}. According to this scenario, the spacecraft starts from $(0,0,0)$ m with zero initial velocities and no attitude with respect to the inertial reference frame. The mission's goal is to maneuver the spacecraft unit to the assembly point within the tight environment created by the target structure components. Consequently, the terminal conditions are the stationary state at the assembly point $(9.5,0,0)$ m, which brings the spacecraft's front to the closest point possible with respect to the end component.

\begin{figure}[h!]
\centering
	\begin{subfigure}{\textwidth}
	\includegraphics[width =\textwidth, height = 10cm]{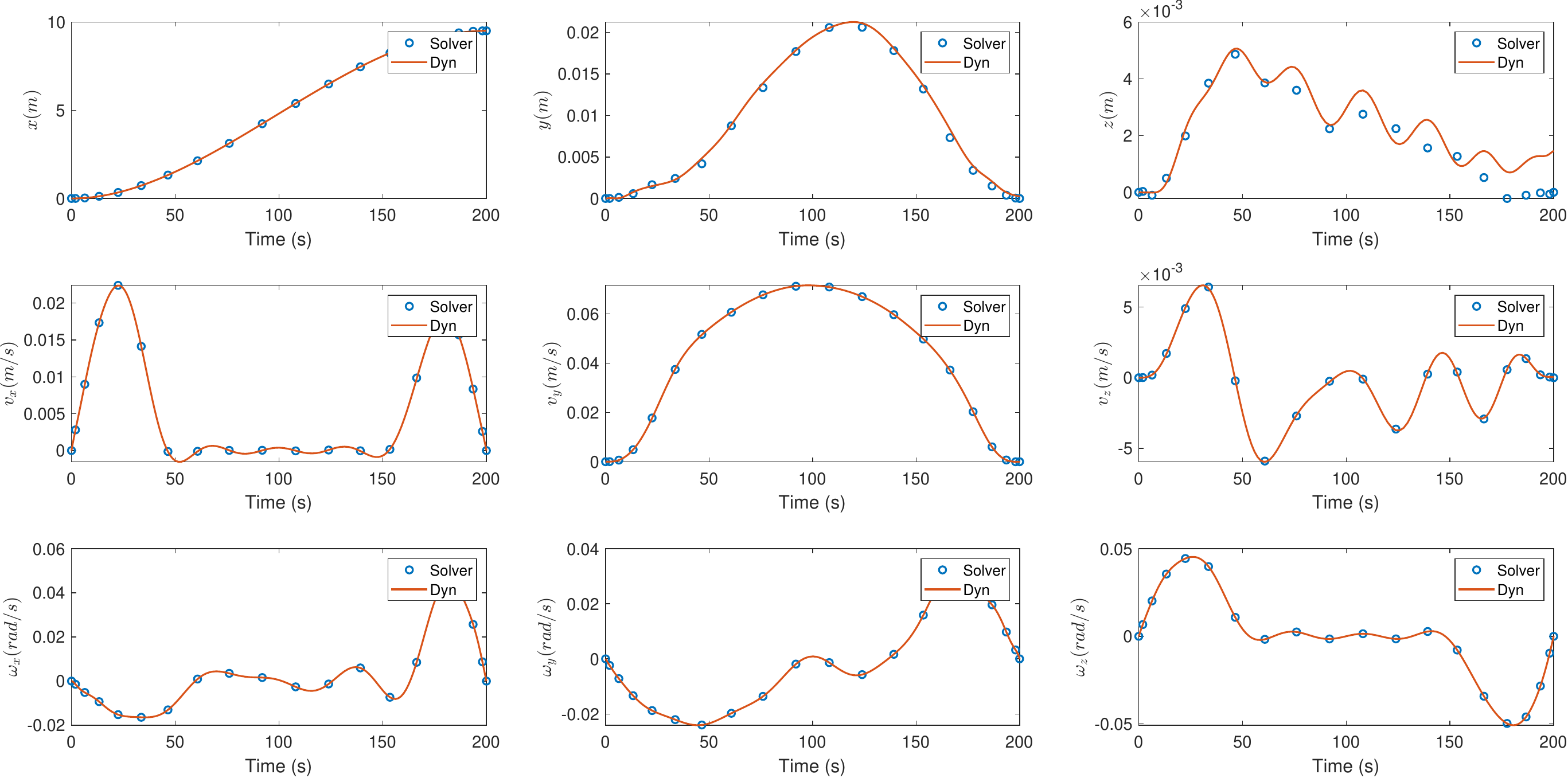}
	\caption{ States: Position (Row 1), Translational Velocity (Row 2) and Angular Velocity (Row 3) }
	\end{subfigure}
	\newline
	\begin{subfigure}{\textwidth}
	\includegraphics[width =\textwidth, height = 7cm]{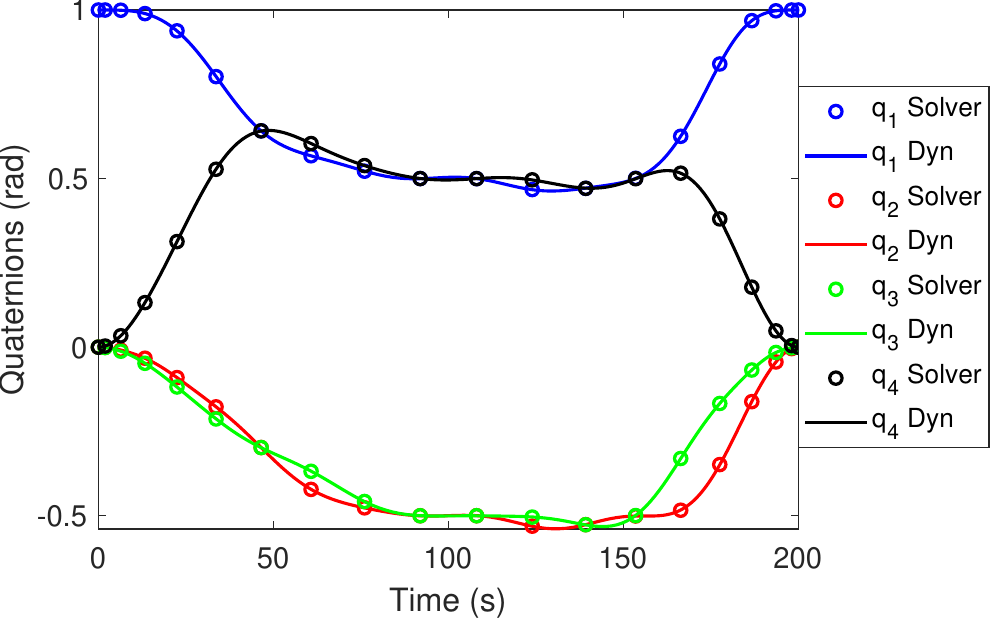}
	\caption{Unit Quaternions}
	\end{subfigure}
\caption{States Time History - Scenario B}
\label{scenB-state}
\end{figure}

\begin{figure}[h]
\centering
\includegraphics[width = \textwidth, height = 10cm]{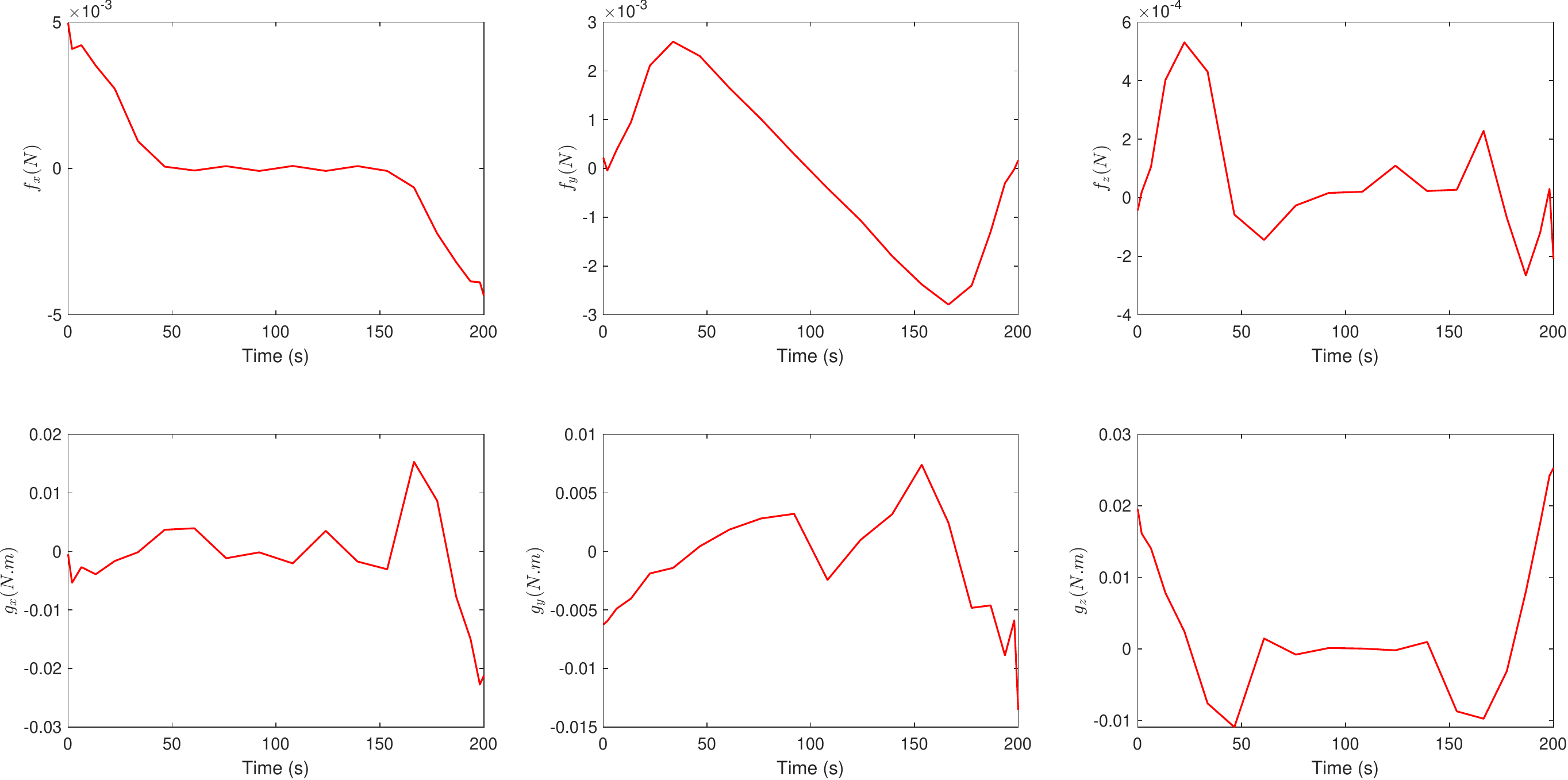}
\caption{Control Inputs Time History - Scenario B}
\label{scenB-controls}
\end{figure}

Again, the optimal control problem, as formulated in this paper, is able to find an optimal trajectory to assemble the desired part, as shown in \figurename \ref{trajB}. The corresponding state and control input time histories are presented in \figurename \ref{scenB-state} and \figurename \ref{scenB-controls} to demonstrate that the generated trajectory is feasible with respect to the state and control input constraints.

The obtained solution from solver was examined against system's dynamics to validate the solver's output, which shows a complete match between the results. Also, the lower Hamiltonian (see \figurename \ref{lowhamB}) provided by the solver's solution confirms that the solution is a Pontryagin's extremal.\\



\subsection{Discussion}\label{discussion}

\begin{figure}[h]
\centering
\includegraphics[width = 0.8\textwidth, height = 8cm]{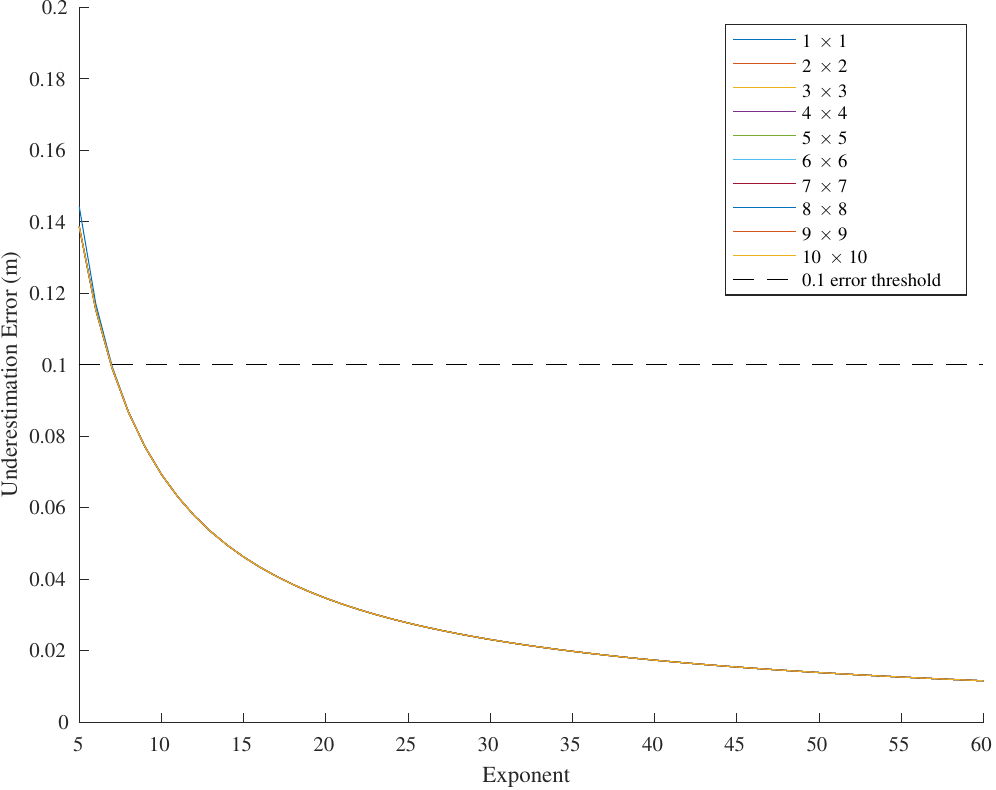}
\caption{Approximation Error for Infinity-norm}
\label{err-analysis}
\end{figure}

In this section, we would like to elaborate on the approximation of the infinity p-norm discussed in Section \ref{sc-st-models}. As illustrated in \figurename \ref{err-analysis}, the underestimation error of the infinity norm is presented for different sizes of a square in a two-dimensional space. The underestimation error decreases as the p-norm increases toward infinity. It is also evident that the size of the polyhedron would not affect the modeling accuracy as shown in \figurename \ref{err-analysis}. As mentioned earlier, this is a sacrifice to lower the dimension of the problem when dealing with several polyhedral objects in an optimal control problem. However, a designer is entirely liberated to directly use the multiple inequalities in Eq. (\ref{d-func}) to precisely model the polyhedral objects without any changes in the differentiable collision avoidance constraints and final results.

The proposed formulation introduces the closest points on the convex sets independent of other optimization parameters. This feature enables the method to handle these points in the desired units different from other parameters' units. To alleviate the approximation error, this feature can be used in the solver in terms of a scaling and balancing algorithm to decrease the object's size \cite{ross2015primer}. Numerical operations can be performed independently of the parameters' units as long as the required transformations are applied when working on shared grounds with other optimization parameters. Therefore, smaller-sized objects can be defined in different units to shrink the  produced values numerically. In this manner, there would be no restriction to use huge exponents which produce numerical complications otherwise. As a result, lower approximation errors can be reached when modeling the polyhedral objects. It should also be noted that carelessly exercising this approach would affect the solution obtained as it could destabilize the solver. 

On another note, an initial guess should be provided to the solver so that it can solve such OCPs. The main reason is that AOA missions are generally complicated, where many variables are involved in the optimization problem. In this paper, we took advantage of a ready-to-use licensed software package \cite{ross2015primer} on MATLAB to solve our problem as a proof of concept of the proposed formulation, which was not appropriate for measuring the computation time due to significant overhead that impacts the solver's performance significantly for large and complicated problems. A solution to this problem would be using lower-level programming languages like Python and Julia, which have very low overheads and plenty of valuable libraries for solving large-scale optimization problems and run much faster than software like MATLAB. To this end, we solved almost the same scenario for another purpose in our other work \cite{tavana2024reinforcement} using Julia programming language, open-source Julia toolbox JuMP \cite{dunning2017jump}, and Interior-point solver IPOPT \cite{wachter2006implementation}. It has been shown that the given OCP takes, on average, 0.1852 seconds to solve when an initial guess is provided, which makes it practical for online motion planning.

To generate an intial guess, the continuation strategy has been employed to build up the complete problem step by step from a trivial problem, and the solution to each sub-problem was used as an initial guess to the next problem which is more complicated and closer to the full problem. To delve deeper, the full problem in the given scenario was broken down into more straightforward problems where the optimal trajectory was generated in the presence of only one component of the target structure in each sub-problem. Afterward, the obtained solutions were superimposed together to form an initial guess for the original problem defined in each scenario. In this manner, the solver quickly converged to an optimal solution, as shown in previous sections. To structure the aforementioned strategy, our other work (see \cite{tavana2024reinforcement} for details) offers a reinforcement learning-based continuation strategy to solve large-scale OCPs, such as the given AOA problem in this paper, without requiring an initial guess.

Ultimately, the technique proposed in this paper can broadly be used in other trajectory generation methods, such as sampling-based methods, like Rapidly exploring Random Tree RRT/$\mbox{RRT}^\star$ \cite{lavalle1998rapidly, faghihi2022kinodynamic}, graph search methods, like $\mbox{A}^\star$ \cite{hart1968formal}, and reachability analysis \cite{margellos2012toward}. The reason is that this formulation establishes the distance between any two objects, modeled by Eqs. (\ref{c-func}) and (\ref{d-func3}), as a step in formulating the collision avoidance constraint. Outside an optimization-based algorithm, this distance can be obtained by solving a system of nonlinear equations as given in Eq. (\ref{prop3}) using fast numerical methods, such as Newton-Raphson method \cite{gil2007numerical}. Therefore, other trajectory generation algorithms, where optimization-based methods are not appropriate, such as high dimensional robotic applications, whether or not considering the system's dynamics, could take advantage of the proposed technique.


\section{Conclusions} \label{conclusions}

This paper presented a novel collision avoidance technique that generates a new set of differentiable constraints in an optimal control problem to guarantee collision-free motion planning, permitting the use of fast gradient-based optimization techniques. We have demonstrated that robotic spacecraft and target structures can be modeled using real-valued convex functions when constructed by the union of a finite collection of convex, compact sets; therefore, the distance between any two sets can be established by extracting the optimality conditions from the corresponding convex optimization problem. The optimal control problem with the new collision avoidance constraints were formulated non-conservatively and are applicable to problems containing point mass and full-dimensional controlled vehicles.

As a numerical case study, an autonomous on-orbit assembly scenario of a complex space structure has been addressed using the new technique. The robotic spacecraft's maneuverability in close proximity of the target space structure using the new collision avoidance constraints has been shown. Also, we have demonstrated in numerical simulations that collision-free assembly operations in tight environments are feasible, and optimal trajectories can be generated for such missions using an optimal control problem. The research continues on tailoring the proposed technique for real-time motion planning and fast computations, as well as adding practical details to the AOA problems for developing robust techniques to perform self-assembly in orbit.\\

\bibliographystyle{elsarticle-num}

\bibliography{TN-Manuscript-Bibliography}

\end{document}